\documentclass[10pt,twocolumn,letterpaper]{article}

\usepackage{iccv}              

\definecolor{iccvblue}{rgb}{0.21,0.49,0.74}
\usepackage[pagebackref,breaklinks,colorlinks,allcolors=iccvblue]{hyperref}

\usepackage{colortbl}
\definecolor{dbcolor}{rgb}{0,0,1}

\usepackage{lscape}

\def\paperID{9584} 
\def\confName{ICCV}
\def\confYear{2025}
\makeatletter
\def\thanks#1{\protected@xdef\@thanks{\@thanks
        \protect\footnotetext{#1}}}
\makeatother

\title{Hybrid-TTA: Continual Test-time Adaptation via\\Dynamic Domain Shift Detection}
\thanks{
This work was supported by IITP grant funded by MSIT (No. RS-2022-00155966: AI Convergence Innovation Human Resources Development (Ewha Womans University) and No. RS-2021-II212068: AI Innovation Hub). $^{\dagger}$ Corresponding author}

\author{
    Hyewon Park \quad\quad\quad
    Hyejin Park \quad\quad\quad
    Jueun Ko \quad\quad\quad
    Dongbo Min$^{\dagger}$\\
    Ewha Womans University, Seoul, South Korea\\
    {\tt\small\{hwpark, clrara, jueun.ko, dbmin\}@ewha.ac.kr}
}

\begin{document}
\maketitle
\begin{abstract}
Continual Test Time Adaptation (CTTA) has emerged as a critical approach to bridge the domain gap between controlled training environments and real-world scenarios. 
Since it is important to balance the trade-off between adaptation and stabilization, many studies have tried to accomplish it by either introducing a regulation to fully trainable models or updating a limited portion of the models.
This paper proposes \textbf{Hybrid-TTA}, a holistic approach that dynamically selects the instance-wise tuning method for optimal adaptation. Our approach introduces Dynamic Domain Shift Detection (DDSD), which identifies domain shifts by leveraging temporal correlations in input sequences, and dynamically switches between Full or Efficient Tuning for effective adaptation toward varying domain shifts. To maintain model stability, Masked Image Modeling Adaptation (MIMA) leverages auxiliary reconstruction task for enhanced generalization and robustness with minimal computational overhead.
Hybrid-TTA achieves $0.6\%p$ gain on the Cityscapes-to-ACDC benchmark dataset for semantic segmentation, surpassing previous state-of-the-art methods. It also delivers about 20-fold increase in FPS compared to the recently proposed fastest methods, offering a robust solution for real-world continual adaptation challenges.

\end{abstract}

\section{Introduction}
\label{sec:intro}

\begin{figure}[!ht]
    \centering
    \includegraphics[width=1.0\columnwidth]{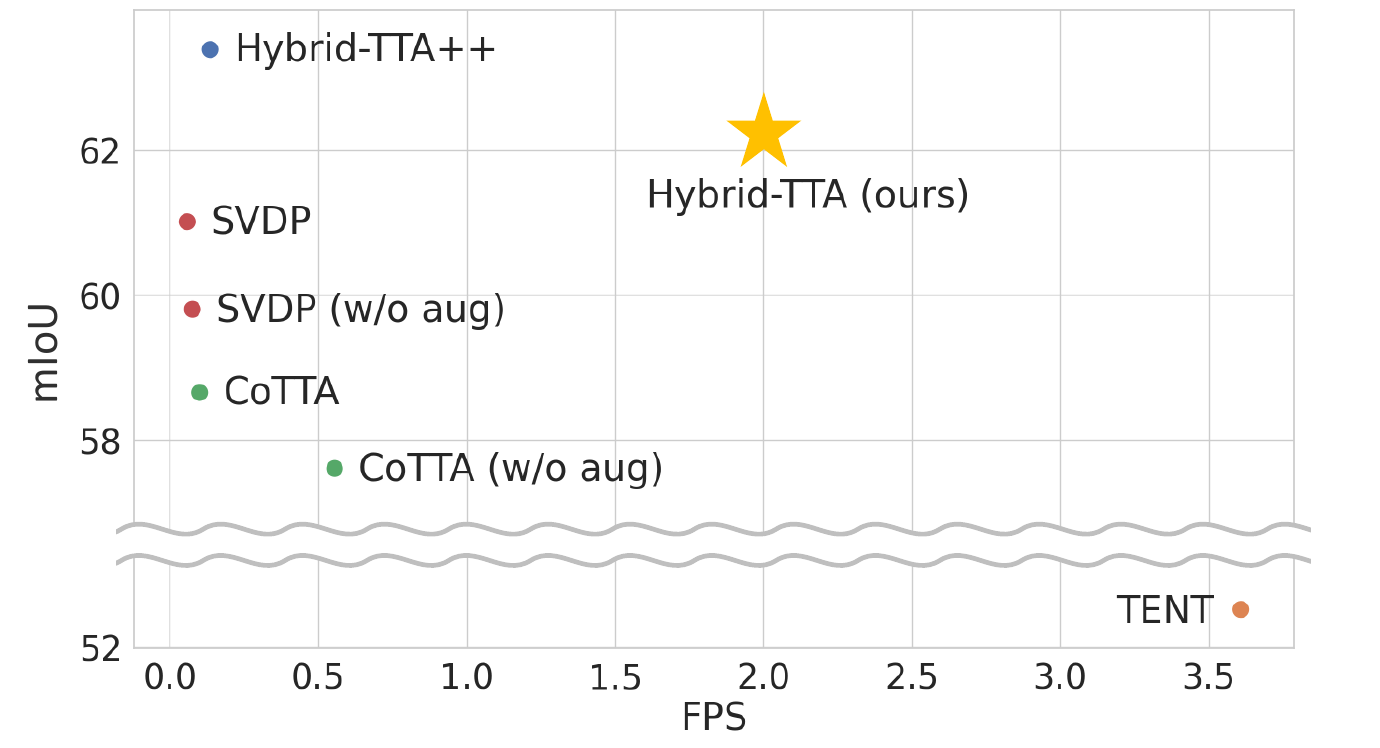}
    \caption{\textbf{Performance vs. FPS on Cityscapes-to-ACDC benchmark.} Hybrid-TTA achieves state-of-the-art 62.2\% mIoU while being significantly faster than other methods with comparable mIoU performance. Hybrid-TTA++ utilizes Test-time Augmentation strategy as well as SVDP and CoTTA, resulting in 63.4\%. See Tab.~\ref{tab:fps}.}
    \label{fig:fps_comparison}
    \vspace{-1em}
\end{figure}


Deep learning has gathered significant attention in computer vision thanks to its powerful representation capability and versatility. Despite achieving high performance on curated datasets, deep learning models face a major challenge: the gap between laboratory-controlled environments and ever-changing real world. These models often struggle to maintain accuracy when faced with diverse and unpredictable realistic scenarios. In this regard, studies including Domain Generalization (DG)~\cite{li2022simple,cho2023promptstyler,tan2024rethinking,ding2022domain,yao2023improving,li2018domain}
and Domain Adaptation (DA)~\cite{hoyer2022daformer,chen2023pipa,hoyer2023mic,hoyer2022hrda,du2024domain,zhang2024universal}
focus on enhancing robustness and adaptability of the model to ensure reliable performance in new, unseen domains.
However, both of them have their own drawbacks: DG, which aims to generalize well to unseen domains, exhibits inherent limitations in adapting to continually changing target environments, as it cannot utilize information from target datasets. DA often struggles with flexibility because it typically assumes a stationary target dataset, limiting its effectiveness in constantly changing target domains.

Given the dynamic nature of target domains during testing, Continual Test-Time Adaptation (CTTA) has emerged as a solution, enabling models to adapt in real-time.
Since CTTA operates in an online fashion, it commonly relies on unsupervised learning methods, such as pseudo-label generation with Mean Teacher~\cite{tarvainen2017mean,wang2022continual}. Unfortunately, unsupervised pseudo labels inevitably contain noise and inaccuracies, leading to error accumulation in the model~\cite{arazo2020pseudo}. Furthermore, as the model adapts to new, unfamiliar target domains, it gradually loses knowledge of the source domain, ultimately leading to catastrophic forgetting and overall performance degradation.

These challenges have spurred the development of recent state-of-the-art methods~\cite{gan2023decorate,liu2023vida,yang2024exploring,yang2024versatile}, attempting to separate domain-agnostic and domain-specific components of a model and focus on training the latter, inspired by Parameter-Efficient Fine-Tuning methods~\cite{chen2022adaptformer,jia2022visual}. On the other hand, other approaches~\cite{wang2022continual,liu2024continual,ma2024improved,lee2024continual} continue to update all parameters, while mitigating these issues by either partially restoring source pre-trained weights or adding regularization term to stabilize the adaptation.


Full Tuning (FT)~\cite{wang2022continual,panagiotakopoulos2022online,liu2024continual,ma2024improved,lee2024continual,liu2023vida,yang2024exploring}, which updates the entire model parameters, is well known for its adaptability, but comes with notable drawbacks. First, \textit{unstable training and error accumulation}, where the reliance on self-generated pseudo labels causes error accumulation and performance degradation. Second, \textit{low efficiency}, as updating the entire model for each target image requires high computational costs.
In contrast, by updating only a subset of the model parameters, Efficient Tuning (ET)~\cite{gan2023decorate,gong2022note,wang2020tent,yang2024versatile} can better preserve the knowledge of the source domain while improving training efficiency.
However, the \textit{limited adaptability} of ET can confine the model to a rudimentary grasp of target domains, leading to suboptimal convergence.
Since the pros and cons of both strategies are in a trade-off, we found that applying the appropriate updating strategy at the right time yields the desired performance, 
balancing plasticity and stability. 
This leads to the important question: \emph{When is the right time to use Full or Efficient Tuning?}


To respond to this perplexing question, we propose \textbf{Hybrid-TTA}, a holistic approach that dynamically alternates between FT and ET, selecting the optimal training strategy for each target instance. 
Hybrid-TTA consists of two synergistic strategies; Dynamic Domain Shift Detection \textbf{(DDSD)} and Masked Image Modeling Adaptation \textbf{(MIMA)}.

\begin{figure}
    \centering
    \includegraphics[width=0.6\columnwidth]{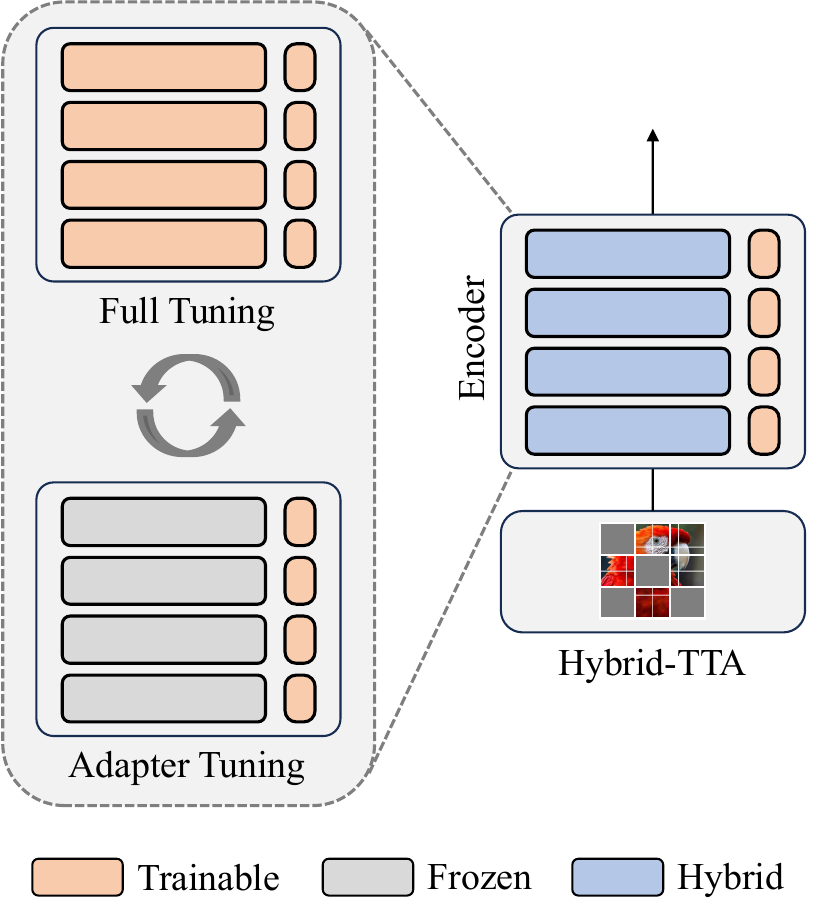}
    \caption{\textbf{Model structure of Hybrid-TTA.} Full Tuning (FT) and Adapter Tuning (AT) are used in hybrid, dynamically alternated fashion by Dynamic Domain Shift Detection (DDSD). When AT is activated, only adapter parameters, implemented via AdaptMLP~\cite{chen2022adaptformer}, are trained.}
    \label{fig:model_structure}
    \vspace{-1.2em}
\end{figure}

Our approach is based on a teacher-student framework, similar to recent CTTA methods~\cite{wang2022continual,liu2023vida}, where the teacher model is updated via an exponential moving average (EMA) of the student model to generate pseudo-labels for the target image.
The DDSD module detects domain shifts 
by capturing a significant drop in temporal correlation among adjacent input images.
Since images from a new domain with different visual features exhibit lower temporal correlation with previous images,
they increase the prediction discrepancy between the pseudo-label of the teacher model and the prediction map of the student model.
This discrepancy upheaval is captured by DDSD based on a \emph{dynamic loss thresholding}, enabling flexible domain shift detection across varying target domains.
If a domain shift is detected, the entire model parameters are updated via FT to maximize the model adaptability toward the domain change. Otherwise, only a small proportion of the parameters are updated via ET to maintain the model stability.
Note that in this paper, we focus on Adapter Tuning (AT)~\cite{chen2022adaptformer} as an exemplification of ET, as depicted in Fig.~\ref{fig:model_structure}.

Furthermore, we propose \textbf{Masked Image Modeling Adaptation (MIMA)} in juxtaposition with DDSD, leveraging the ability of Masked Image Modeling (MIM) for continual model adaptation.
MIM~\cite{xie2022simmim,he2022masked} is well-known for its robustness and generalizability as a self-supervised learning method and has been applied for model adaptation~\cite{hoyer2023mic,liu2024continual,gandelsman2022test}.
Prior work~\cite{hoyer2023mic} proposed Masked Image Consistency (MIC), an Unsupervised Domain Adaptation (UDA) method that uses masked images as input for the student model, to enhance the learning of contextual relationships and better utilize contextual cues in the target domain.
However, the fundamental characteristics of UDA — accessible source dataset, mini-batch training, and repeated training on target dataset
— 
favor MIC, making its direct application to CTTA challenging and necessitating an optimized approach.

Our simple yet effective CTTA-tailored MIM method, MIMA, integrates image reconstruction as an auxiliary task alongside the primary task of semantic segmentation. While providing the model with randomly masked input images as MIC, MIMA extends its capability by training on both tasks simultaneously, encouraging both the primary segmentation loss and the reconstruction loss to contribute to adaptation.
This dual objective approach leverages the self-supervising nature of MIM for stable pseudo-label generation, and maximizes the use of image-driven information from the original target image. 
It is noteworthy that while recent CTTA methods \cite{wang2022continual,liu2023vida, yang2024exploring} rely on Test-time Augmentation for a stable adaptation at the cost of low frame per second (FPS) (See Tab.~\ref{tab:fps}), our method does not employ any additional pseudo-label enhancement techniques while delivering the state-of-the-art performance.

\noindent We summarize our main contributions as follows:
\begin{itemize}
    \item We introduce a novel CTTA framework that alternates ET and AT for instance-wise model adaptation, balancing the trade-off between model adaptability and stability for dynamic target domains.    
    \item To determine the training strategy for each target instance, we propose Dynamic Domain Shift Detection (DDSD) based on dynamic loss thresholding, which selects the optimal tuning method for each target instance.
    \item We introduce Masked Image Modeling Adaptation (MIMA), which integrates self-supervised image reconstruction as an auxiliary task to maximize learning from target images, leading to improved model stability and robustness.
    \item Our approach outperforms state-of-the-art methods on multiple CTTA benchmark datasets for semantic segmentation, while significantly reducing inference time.
\end{itemize}

\section{Related Work}

\noindent\textbf{Continual Test-Time Adaptation}  CTTA addresses the challenge of adapting models in dynamic environments where target domains change over time, while traditional Test-Time Adaptation (TTA) methods assume a static target domain~\cite{yang2021generalized,lim2023ttn,kundu2020universal}. Recent CTTA works have employed techniques such as self-training and entropy minimization to adapt the model online~\cite{wang2022continual,wang2020tent}. \cite{wang2022continual} mitigates error accumulation and catastrophic forgetting using Test-time Augmentation and stochastic weight reset, fully updating the source model for continual target domains.

Several parameter-efficient CTTA methods aim to mitigate catastrophic forgetting while reducing computational overhead. For instance, \cite{song2023ecotta} employs a meta-network for stability, while \cite{gan2023decorate,gao2022visual} update a small subset of parameters named visual domain prompts to improve efficiency. \cite{liu2023vida} leverages multi-rank adapters to capture both domain-agnostic and -specific features. Meanwhile, \cite{liu2024continual} employs HOG-feature reconstruction for robust adaptation, though its reliance on a handcrafted module may limit overall performance.
\cite{wang2020tent} minimizes prediction entropy by adapting the batch normalization parameters, but struggles due to its limited adaptability.

\noindent\textbf{Masked Image Modeling (MIM)}  MIM has gained prominence in self-supervised learning, enabling models to learn robust visual representations by reconstructing masked portions of images. Notable examples include \cite{he2022masked} and \cite{xie2022simmim} which mask significant parts of an image and train the model to reconstruct them, learning high-quality feature representations that benefit various downstream tasks.

Building on these principles, \cite{hoyer2023mic} uses MIM in unsupervised domain adaptation (UDA) for semantic segmentation by minimizing the consistency loss between teacher's pseudo-label from the original image and student's prediction map from the masked image. Recent works also incorporate MIM into Test-Time Training (TTT), such as \cite{gandelsman2022test} using MIM for both source training and test-time feature refinement in TTT, improving its adaptability to continual domain shifts. Also, \cite{mansour2024ttt} was proposed which leverages the model's ability of reconstruction to generate pseudo-labels for the main task.


\noindent\textbf{Domain Shift Detection for CTTA}  Recent studies including \cite{shi2024controllable} attempt to quantify distributional changes in input dataset via uncertainty estimation and statistical divergence metrics.
In contrast, other techniques~\cite{colomer2023adapt,panagiotakopoulos2022online} detect domain shifts by analyzing model responses through feature distance or confidence measurements.
Our work, building on these ideas, determines domain shifts by measuring the difference in model responses between the student and temporally correlated teacher networks.



\begin{figure*}[t!]
\centering
\includegraphics[width=2.1\columnwidth]{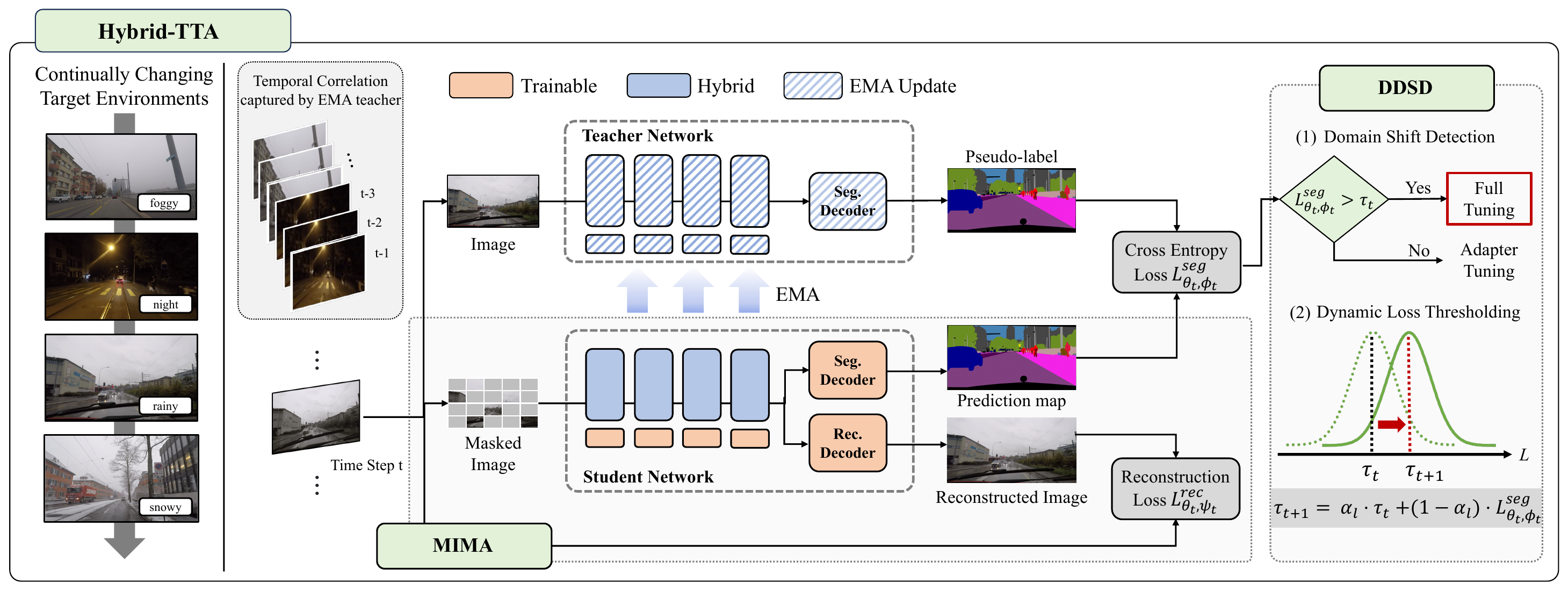}
\caption{\textbf{Overview of the Hybrid-TTA framework.}
Continual Test-time Adaptation with \textbf{Masked Image Modeling Adaptation (MIMA)} under continually changing target environments (e.g., foggy, night, rainy, snowy conditions). Here, \textbf{Dynamic Domain Shift Detection (DDSD)} is used along with MIMA, which detects the domain shift and switches between Full Tuning and Adapter Tuning via Dynamic Loss Thresholding to maintain optimal adaptation performance.}
\label{fig:2-Hybrid-TTA}
\end{figure*}

\section{Preliminary}
\label{sec:preliminary}

\paragraph{Continual Test-time Adaptation (CTTA)}
CTTA aims to adapt a model, which is initially trained on source data $\mathcal{D}_s = (\mathcal{X}_s, \mathcal{Y}_s)$, to multiple unlabeled target data \(\mathcal{D}_{T} = \{\mathcal{X}_{T_1}, \mathcal{X}_{T_2}, \dots, \mathcal{X}_{T_n}\}\) during deployment, where $n$ represents the number of unseen domains.
The entire adaptation process is constrained by two key challenges: (1) the model cannot access any source data during test-time adaptation, and (2) the model can only access each instance of target domain once, necessitating efficient and effective use of respective target instance.
When the target domain data $\mathcal{X}_{T_i}$ consists of $N_i$ target samples for $i=1,\dots,n$, 
for a simplicity of notations, we denote $x_t$ by the $t^{th}$ target instance for $t=0,\dots,|\mathcal{D}_{T}|-1$, where $|\mathcal{D}_{T}|=\sum\nolimits_{i = 1}^n {N_i}$.
Similarly, in the following sections of this paper, we omit the domain notation $T_1,...,T_n$ as all target domains are considered to be integrated into a single sequence.

To tackle the challenges of CTTA as an unsupervised online learning, a common strategy is to use a weight-averaged teacher model to generate pseudo-labels for the target data~\cite{tarvainen2017mean,wang2022continual}. In source training, the student encoder $f_\theta$ and task decoder $h_\phi$ is trained on the source dataset $\mathcal{D}_{s}$, establishing the source-trained parameters $\theta_{s}, \phi_{s}$.
During test-time, the teacher model $f'_{\theta}, h'_{\phi}$ is firstly initialized with these parameters $\theta_{s}, \phi_{s}$.
As the model encounters new target data over time $t$, the teacher model generates pseudo-labels $\hat{y}'_{t} = h_{\phi'_t}(f_{\theta'_t}(x_t))$ for each input $x_t$. 
The student model is then updated by minimizing the task loss $\mathcal{L}^{seg}_{\theta_t, \phi_t}$ between its own prediction $\hat{y}_t$ and the teacher's pseudo-label $\hat{y}'_t$.

After each update of the student model, the weights of the teacher model $\theta', \phi'$ are adjusted using an exponential moving average (EMA) of the student's updated weights:
\begin{equation}\label{eqn:3.1.1}
\theta'_{t+1} = \alpha \cdot \theta'_t + (1-\alpha) \cdot \theta_{t+1}
\end{equation}
where $\alpha$ is a smoothing factor that governs the influence of the student's weights on the teacher model's updates. \eqref{eqn:3.1.1} applies for both $\theta'$ and $\phi'$.



\paragraph{Masked Image Modeling (MIM)}
MIM is broadly utilized for effective pretraining~\cite{xie2022simmim, he2022masked}. Input images are partially occluded with random patch masks and passed through the Transformer encoder, and the original intensities of the masked patches are reconstructed in a decoder. The training objective is typically set as:
\begin{equation}
\begin{aligned}
\min_{\theta, \psi} \mathop{\mathbb{E}}_{x \sim \mathcal{D}}
&\mathcal{L}_{rec} (x\odot M,g_\psi(z)\odot M) ,
\quad z = f_\theta (\Tilde{x}), \\
&\Tilde{x} = x \odot (1 - M) + e\odot M
\end{aligned}
\label{eqn:3.4.1}
\end{equation}

\noindent where $\odot$ represents an element-wise multiplication, and $\mathcal{D}$ is the train data distribution. $M$ denotes a patch mask, whose element is 1 if masked, 0 otherwise\footnote{$x \odot (1-M)$ represents ``visible patches" and vice versa.}. $f(\cdot)$ and $g(\cdot)$ are encoder and reconstruction decoder respectively, with parameters $\theta$ and $\psi$, and $z$ is the learned representation. The masked input $\Tilde{x}$ is generated by replacing the masked patches in $x$ with trainable mask tokens $e$. The reconstruction loss $\mathcal{L}_{rec}$ is computed only for masked regions.


\section{Proposed Method}
We introduce \textbf{Hybrid-TTA}, a specially designed CTTA framework, illustrated in Fig.~\ref{fig:2-Hybrid-TTA}. To effectively manage the wide range of domain gaps presented by continually changing domains, a CTTA model must achieve the dual objectives of adaptability and stability, two distinctive qualities in a trade-off.
Since FT offers flexible adaptation but struggles with instability, while AT is more reliable and efficient but has limited adaptability, Hybrid-TTA aims to accomplish both goals by dynamically alternating these adaptation methods through DDSD, as well as fully exploiting useful image features through complementary application of segmentation and reconstruction tasks via MIMA.

\subsection{Dynamic Domain Shift Detection (DDSD)}
\label{sec:ddsd}


As previously discussed, correctly determining when to use FT or AT is crucial for maximizing the benefits of Hybrid-TTA.
Given that images from new, unseen domains tend to require adaptability and those from familiar domains prioritize stability, \emph{domain shifts} can serve as a key indicator for alternating FT and AT.
However, detecting domain shift is a challenging task due to the vast number of possibilities in real-world environments. 
{
Dynamic Domain Shift Detection (DDSD) aims to solve the problem using two distinctive approaches, temporal correlation and dynamic loss thresholding.

\subsubsection{Domain Shift Detection via Temporal Correlation}

In continually changing environment, a domain shift refers to a change in visual signals over time, such as an abrupt change in weather. These shifts cause distributional changes in the input image sequence, consequently altering the model's predictions over time.
This can be understood through the concept of \emph{temporal correlation}, which captures the consistency in model responses across consecutive time steps. In short, a domain shift causes the model to respond differently, reducing the temporal correlation.



To leverage this concept for domain shift detection,
it is necessary to maintain the history of previously seen images to compare with current input.
Previous studies often utilize a replay buffer~\cite{cho2023complementary} to retain historical information by sampling and storing a subset of prior inputs, which requires additional memory.
In contrast, we argue that the Mean Teacher model~\cite{tarvainen2017mean} can be viewed as an alternative mechanism for maintaining past domain characteristics, thanks to its evolving nature via EMA.
Unlike the student model, which undergoes rapid updates, the teacher model evolves gradually, allowing it to retain temporal correlations in its predictions.
So, if a domain shift occurs, the teacher model's predictions would differ from those of the student model, lowering the temporal correlation.

DDSD captures this discrepancy by comparing the student's predicted segmentation map and the teacher's pseudo-label using the cross-entropy loss. It effectively quantifies the divergence between dense segmentation maps while avoiding additional computational overhead, making DDSD exceptionally efficient for detecting domain shifts.
Using this information, DDSD dynamically selects the optimal tuning strategy for each instance: FT is applied when a significant loss upheaval (\emph{i.e.}, low correlation) indicates a major domain shift;
AT is used when the loss remains relatively unchanged (\emph{i.e.}, high correlation), implying the domain is stable.

}

\subsubsection{Dynamic Loss Thresholding}

Detecting domain shifts through temporal correlation often relies on a fixed threshold. Yet, this approach lacks flexibility, especially in CTTA scenarios where the model is continually evolving and the degree of domain shift varies across target instances.



To overcome this limitation, we introduce \emph{Dynamic Loss Thresholding}, where the loss threshold evolves over time in addition to model adaptation.
Assume $\mathcal{L}^{seg}_{\theta_t,\phi_t}$ is a segmentation loss, calculated between the teacher's pseudo label and the student's prediction map at time step $t$.
An initial loss threshold $\tau_0$ is set to 0, under the assumption that all upcoming target instances are under domain shift. At each time step $t$, this threshold is updated using an EMA of the newly calculated loss:
\begin{equation}\label{eqn:3.3.2}
 \tau_{t+1} = \alpha_l \cdot \tau_t + (1 - \alpha_l) \cdot \mathcal{L}^{seg}_{\theta_t, \phi_t}
\end{equation}
where $\alpha_l$ is a smoothing factor for the EMA of loss threshold with default value 0.999~\cite{tarvainen2017mean}.




Hence, if a loss value at time $t$ exceeds the threshold, \emph{i.e.} $\mathcal{L}^{seg}_{\theta_t,\phi_t} > \lambda_{d}\cdot\tau_t$, DDSD identifies a domain shift, assuming that the temporal discrepancy between the student and teacher model is significant. Note that $\lambda_{d}$ is sensitivity hyperparameter (See supplementary material).


\subsection{Masked Image Modeling Adaptation (MIMA)}

As previously mentioned, the critical challenge of CTTA is that the target images arrive without ground truth labels. This makes adaptation particularly difficult, as continually changing target domains often cause unstable adaptation.
To mitigate error accumulation caused by 
pseudo-labels and fully exploit the visual information provided by target images, we incorporate a self-supervised learning method, Masked Image Modeling (MIM), to our CTTA framework.


As illustrated in Fig.~\ref{fig:2-Hybrid-TTA}, the target input image is masked and fed into the feature extraction encoder of the student model to generate a masked image representation. This representation is subsequently passed through two distinctive decoders for segmentation and image reconstruction, producing a prediction map and a reconstructed image, respectively.
The segmentation loss is computed between the teacher's pseudo-labels and the student's predictions as following:
\begin{equation}\label{eqn:3.3.3}
    \mathcal{L}^{seg}_{\theta_t, \phi_t} = CE  (\hat{y}'_t, h_{\phi_t}(f_{\theta_t}(\Tilde{x}_t))), 
\end{equation}
where $\hat{y}'_t = h_{\phi'_t}(f_{\theta'_t}(x_t))$ represents the pseudo-label generated by the teacher model.
For image reconstruction, the model aims to reconstruct the original image from the masked input $\Tilde{x}$ based on \eqref{eqn:3.4.1}. At time step $t$, the reconstruction loss between the original and reconstructed images is calculated using $L1$ loss, as below:
\begin{equation}\label{eqn:3.4.2}
    \mathcal{L}^{rec}_{\theta,\psi} = \frac{1}{|M|}\parallel (x - g_\psi(f_\theta(\Tilde{x})))\odot M\parallel_1,
\end{equation}

A final loss is computed as in \eqref{eqn:3.3.2} and updated at each time $t$, as the model continuously adapts during the TTA process.
\begin{equation}\label{eqn:3.3.2}
    \mathcal{L}_{total} = \mathcal{L}^{seg}_{\theta,\phi} + \lambda_{r} \cdot \mathcal{L}^{rec}_{\theta,\psi}
\end{equation}
Note that the reconstruction decoder is randomly initialized, not requiring pre-training or warm-up stage.
Further experiments with respect to the reconstruction loss weight $\lambda_{r}$ are provided in the supplementary material.

It is noteworthy that Masked Image Consistency (MIC) was proposed in \cite{hoyer2023mic} as a UDA plug-in that trains the student network to predict the complete segmentation map from randomly masked images to enhance contextual understanding of the target domain.
However, key characteristics of UDA — access to the source dataset, mini-batch training, and repeated training on target dataset — enable MIC to maintain model stability while fully leveraging its potential. In contrast, the absence of these characteristics in CTTA makes models vulnerable to error accumulation, leading to unstable pseudo-labels.
To fully exploit the useful features of the target image in a challenging CTTA setup where adaptation is conducted once and only with a single target image, we introduce MIMA, a MIM-based strategy that integrates masked image reconstruction and masked image segmentation, enhancing generalization and robustness as validated in experiments of supplementary material. 

\section{Experiments}
\label{sec:experiments}

\begin{table*}[ht!]
\scriptsize\addtolength{\tabcolsep}{-2.2pt}
\centering
\caption{Comparison of CTTA segmentation performance on \textbf{Cityscapes-to-ACDC benchmark} over 10 rounds. Mean is the average score of mIoU. The best results are highlighted in \textbf{bold}. Only 1, 4, 7 and 10th round scores are written, while the entire scores are provided in suppl. material.
Scores of additional methods including VDP~\cite{gan2023decorate} and C-MAE~\cite{liu2024continual} can also be found in supplementary material.
}
\label{tab:performance_comparison}
\begin{tabular}{lllllllll|lllllllll}
 & \multicolumn{16}{c}{\hspace{2cm} \textbf{t} $\xrightarrow{\hspace{12cm}}$}& \\
\end{tabular}
\begin{tabular}{l|llll|llll|llll|llll|l}
\toprule
\bf{Test} & \multicolumn{4}{c}{1}& \multicolumn{4}{c}{4}& \multicolumn{4}{c}{7}& \multicolumn{4}{c|}{10} & All$\uparrow$ \\
\bf{Condition} & \textbf{Fog} & \textbf{Night}  & \textbf{Rain} & \textbf{Snow}  & \textbf{Fog} & \textbf{Night} & \textbf{Rain} & \textbf{Snow} & \textbf{Fog} & \textbf{Night} & \textbf{Rain} & \textbf{Snow} & \textbf{Fog} & \textbf{Night}  & \textbf{Rain} &\textbf{Snow}  & Mean \\
\midrule
(a) Source
    & 69.1 & 40.3 & 59.7 & 57.8 
    & 69.1 & 40.3 & 59.7 & 57.8
    & 69.1 & 40.3 & 59.7 & 57.8
    & 69.1 & 40.3 & 59.7 & 57.8
    & 56.7 \\
(b) BN Stats Adapt \cite{schneider2020improving}
    & 62.3 & 38.0 & 54.6 & 53.0
    & 62.3 & 38.0 & 54.6 & 53.0
    & 62.3 & 38.0 & 54.6 & 53.0
    & 62.3 & 38.0 & 54.6 & 53.0
    & 52.0 \\
(c) TENT-continual \cite{wang2020tent}
    & 69.0 & 40.2 & 60.1 & 57.3
    & 66.5 & 36.3 & 58.7 & 55.0
    & 64.2 & 32.8 & 55.3 & 50.9
    & 61.8 & 29.8 & 51.9 & 47.8
    & 52.3 \\
(d) CoTTA \cite{wang2022continual}
    & 70.9 & 41.2 & 62.4 & 59.7
    & 70.9 & 41.0 & 62.7 & 59.7
    & 70.9 & 41.0 & 62.8 & 59.7
    & 70.9 & 41.0 & 62.8 & 59.7
    & 58.6 \\
(e) EcoTTA \cite{song2023ecotta}
    & 68.5 & 35.8 & 62.1 & 57.4
    & 68.1 & 35.3 & 62.3 & 57.3
    & 67.2 & 34.2 & 62.0 & 56.9
    & 66.4 & 33.2 & 61.3 & 56.3
    & 55.2 \\
(f) SVDP \cite{yang2024exploring}
    & 71.0 & 43.2 & 64.8 & 62.0
    & \bf{71.8} & 43.9 & 66.6 & 62.3
    & 71.9 & 43.6 & 66.5 & 62.3
    & 71.6 & 43.5 & \bf{66.6} & 62.0
    & 61.0 \\
(g) ViDA \cite{liu2023vida}
    & \textbf{71.6} & 43.2 & \bf{66.0} & \textbf{63.4}
    & 70.9 & 44.0 & 66.0 & \bf{63.2}
    & \textbf{72.3} & 44.8 & 66.5 & 62.9
    & \textbf{72.2} & 45.2 & 66.5 & 62.9
    & 61.6 \\
\midrule
\bf{(h) Hybrid-TTA}
    & 70.3 & \bf{44.5} & 65.1 & 63.2
    & 70.1 & \bf{49.3} & \bf{66.9} & 63.1
    & 69.6 & \bf{49.4} & \bf{66.7} &\bf{63.0}
    & 69.9 & \bf{49.5} & 66.5 & \bf{63.0}
    & \bf{62.2} \\
\bottomrule
\end{tabular}
\vspace{-0.2cm}
\end{table*}

\begin{figure*}[!ht]
    \centering
    \includegraphics[width=0.8\paperwidth]{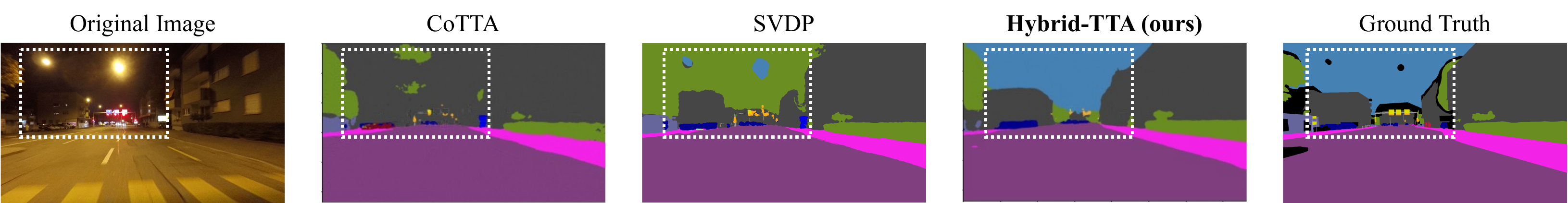}
    \caption{\textbf{Qualitative comparison on the Night domain of Cityscapes-to-ACDC benchmark.}}
    \label{fig:qualitative comparison}
    \vspace{-0.5cm}
\end{figure*}

\begin{figure}[t]
    \centering
    \includegraphics[width=0.85\columnwidth]{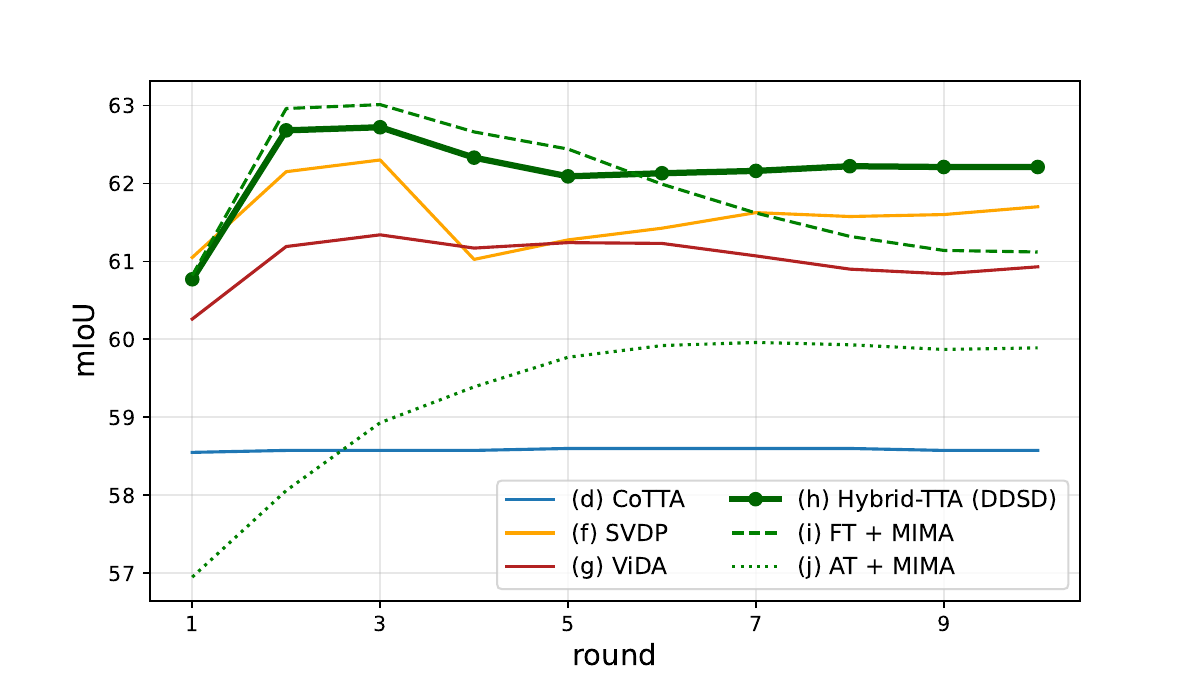}
    \vspace{-0.2cm}
    \caption{{\bf Performance of various CTTA methods on Cityscapes-to-ACDC benchmark.}
    }
    \label{fig:4.1-performance graph}
    \vspace{-0.6cm}
\end{figure}

\subsection{Experimental Setups}

\noindent\textbf{Dataset and Task Setup}
Experiments are carried out on the Cityscapes-to-ACDC~\cite{cordts2016cityscapes,sakaridis2021acdc} benchmark, which assesses test-time performance of CTTA methods under four weather conditions (fog, night, rain, and snow) repeated cyclically for 10 times to evaluate long-term adaptation.
We down-sampled Cityscapes to 1024$\times$1024 and ACDC to 960$\times$544, as the image sizes need to be multiple of the mask patch size 32.
We also follow the experimental setup of \cite{panagiotakopoulos2022online} for assessment on the OnDA benchmark, which applies synthetic rain~\cite{tremblay2021rain} of various rain intensities (25mm-200mm) to the Cityscapes~\cite{cordts2016cityscapes} dataset. Note that the five domains in OnDA benchmark are cyclically repeated 5 times. 
We adopt two types of benchmarks to show the model performance in different environments, considering that OnDA benchmark contains a smaller domain gap with the source dataset (Cityscapes) compared to ACDC.


\noindent\textbf{Implementation Details}
We follow the setup of the previous work~\cite{wang2022continual}, 
using SegFormer MiT-B5~\cite{xie2021segformer} trained on down-sampled Cityscapes (1024$\times$1024).
The same pre-trained weights are used for initialization by all participating methods for a fair comparison.
To implement Adapter Tuning, we incorporate AdaptMLP~\cite{chen2022adaptformer} in each transformer layer, which adds 0.17M extra parameters to the model, occupying 0.1\% out of total model parameters.
The adapter parameters are initialized according to \cite{chen2022adaptformer} without any warm-up training.
All experiments are conducted using NVIDIA A6000-ada and RTX3090. Further details are in the supplementary material.


\begin{table*}[ht!]
\scriptsize\addtolength{\tabcolsep}{-2.2pt}
\centering
\caption{Comparison of CTTA segmentation performance on \textbf{OnDA benchmark} over 5 rounds. Mean is the average score of mIoU. The best results are highlighted in \textbf{bold}. Only 1, 3 and 5th round scores are written, while the entire scores are provided in suppl. material.}
\vspace{-1em}
\label{tab:performance_comparison_onda}
\begin{tabular}{lllllllll|lllllllll}
 & \multicolumn{16}{c}{\hspace{2.2cm} \textbf{t} $\xrightarrow{\hspace{12cm}}$}& \\
\end{tabular}
\begin{tabular}{l|cccccc|cccccc|cccccc|l}
\toprule
\bf{Test} & \multicolumn{6}{c}{1} & \multicolumn{6}{c}{3} & \multicolumn{6}{c|}{5} & All$\uparrow$ \\
\bf{Intensity (mm)}
    & \textbf{25} & \textbf{50} & \textbf{75} & \textbf{100} & \textbf{200} & \textbf{mean}
    & \textbf{25} & \textbf{50} & \textbf{75} & \textbf{100} & \textbf{200} & \textbf{mean}
    & \textbf{25} & \textbf{50} & \textbf{75} & \textbf{100} & \textbf{200} & \textbf{mean} & Mean \\
\midrule
(a) Source
    & 67.7 & 65.6 & 62.9 & 58.1 & 45.3 & 59.9
    & 67.7 & 65.6 & 62.9 & 58.1 & 45.3 & 59.9
    & 67.7 & 65.6 & 62.9 & 58.1 & 45.3 & 59.9
    & 62.4 \\
(b) TENT-continual \cite{wang2020tent}
    & 66.8 & 64.8 & 62.3 & 58.1 & 45.6 & 59.5
    & 64.2 & 61.3 & 58.6 & 54.6 & 43.1 & 56.4
    & 61.1 & 58.1 & 55.4 & 51.8 & 41.2 & 53.5
    & 56.5 \\
(c) CoTTA \cite{wang2022continual}
    & 68.9 & 67.6 & 66.8 & 65.6 & 59.2 & 65.6
    & \bf{68.7} & 67.5 & 66.6 & 65.6 & 60.1 & 65.7
    & \bf{68.6} & \bf{67.6} & 66.7 & 65.7 & 60.7 & 65.9
    & 65.8 \\
(d) SVDP \cite{yang2024exploring}
    & \bf{69.2} & \bf{68.1} & 66.9 & 65.5 & 61.7 & 66.3
    & 67.5 & 66.8 & 66.1 & 65.3 & 62.6 & 65.7
    & 67.4 & 66.6 & 66.1 & 65.4 & 62.1 & 65.5
    & 65.7 \\
\midrule
\textbf{(e) Hybrid-TTA}
    & 68.2 & 67.7 & \bf{67.4} & \bf{66.8} & \bf{62.9} & 66.6
    & 68.1 & \bf{67.7} & \bf{67.1} & \bf{66.5} & \bf{64.5} & 66.8
    & 67.4 & 67.1 & \bf{66.7} & \bf{66.0} & \bf{64.3} & 66.3
    & \bf{66.7} \\
\bottomrule
\end{tabular}
\vspace{-0.2cm}
\end{table*}

\subsection{Main Results}
\label{sec:main_results}

\noindent\textbf{Cityscapes-to-ACDC benchmark}
Tab.~\ref{tab:performance_comparison} provides a comparative study with various CTTA methods on the Cityscapes-to-ACDC benchmark. Here, `Mean' represents an average mIoU score over 10 rounds of experiments. Fig.~\ref{fig:4.1-performance graph} visualizes the 10-round scores of some methods. Note that (c) TENT-continual is a variant of TENT~\cite{wang2020tent} originally proposed for test-time adaptation, with a slight modification to continuously update batch normalization parameters using entropy loss to make it suitable for CTTA.
A source-trained model (a) SegFormer MiT-B5 works as a baseline with no adaptation. The performance of (b) BN Stats Adapt~\cite{schneider2020improving} and (c) TENT-continual~\cite{wang2020tent} gradually declines due to catastrophic forgetting and limited adaptability. (e) EcoTTA~\cite{song2023ecotta}, a CNN based ET method, shows a similar performance degradation. (d) CoTTA~\cite{wang2022continual} maintains stable performance by preventing error accumulation with test-time augmentation, but fails to effectively adapt to the target domain, resulting in only moderate gains.

Current state-of-the-art methods, (f) SVDP~\cite{yang2024exploring} and (g) ViDA~\cite{liu2023vida}, achieve a significant performance gain compared to previous methods (a-e) by more than 2.4$\%p$.
Our approach, (h) Hybrid-TTA, shows outstanding results compared to (f) and (g). In particular, as Fig.~\ref{fig:4.1-performance graph} shows, our method achieves a large performance leap in the first couple of rounds, as the DDSD module in Hybrid-TTA prefers to choose FT in the early stages of adaptation to overcome the initial domain gap between the source and target domains. We notice that the performance remains fairly stable over the remaining 8 rounds, where DDSD mostly selects AT to ensure stable adaptation and computational efficiency. This phenomenon is discussed in detail in Sec.~\ref{sec:ddsd-add-behaviour}.

In addition, Fig.~\ref{fig:4.1-performance graph} validates the effectiveness of DDSD of Hybrid-TTA by comparing it with FT and AT. In this graph, (h) Hybrid-TTA (DDSD) represents the result of Tab.~\ref{tab:performance_comparison}. (i) always applies FT throughout all rounds, while (j) uses only AT. Note that MIMA is used in all cases for a fair comparison. The graph shows that (i), despite achieving good performance in the first three rounds, rapidly declines due to error accumulation and catastrophic forgetting caused by excessive updates, resulting in average performance of 61.9\%. On the other hand, (i) exhibits suboptimal performance due to its limited adaptability, resulting in 59.27\%. In contrast, (h), where DDSD orchestrates the adaptation, effectively leverages the strengths of both strategies, leading to outstanding results of 62.2\%.

Furthermore, our approach shows a significant improvement in the Night domain (44.5\%$\rightarrow$49.5\%) of Tab.~\ref{tab:performance_comparison}, where (f) and (g) shows relatively moderate gains (43.2\%$\rightarrow$43.5\%, 43.2\%$\rightarrow$45.2\%). This is impressive because the Night domain is the most challenging with the largest domain gap from the source dataset due to the poor lighting condition~\cite{liu2023vida}.
Fig.~\ref{fig:qualitative comparison} confirms better visual results of our method in the Night domain. Further qualitative comparisons are provided in the supplementary material.

\noindent\textbf{OnDA benchmark} Tab.~\ref{tab:performance_comparison_onda} presents comparative results on the OnDA benchmark.
(a) Source and (b) TENT-continual show a significant performance drop from easy (25mm) to hard (200mm) domains (\emph{e.g.,} 67.7\%$\rightarrow$45.3\%), due to limited adaptation capability. Moreover, the average performance of (b) per round severely declines due to catastrophic forgetting. (c) CoTTA shows fine mean mIoU (65.8\%) thanks to its increased learning capacity and robust pseudo-labels. (c) also maintains steady performance, even slightly improving in the hard domain of 200mm (59.2\%$\rightarrow$60.7\%), but still experiences severe drop from easy to hard domains ($-9.7\%p$) in the $1^{st}$ round. 
(d) SVDP reduces the impact of the domain shift, as the score gap from easy to hard domains in the $1^{st}$ round is reduced ((c) $-9.7\%p$$\rightarrow$(d) $-7.5\%p$), but still shows moderate performance in the hard domain (61.7\%$\rightarrow$62.1\%).

In contrast, (e) Hybrid-TTA achieves a total mean of 66.7\%, outperforming all methods by approximately $1\%p$. It also shows significant improvement in 200mm domain (62.9\%$\rightarrow$64.3\%) due to its improved adaptability. Moreover, (e)'s mean mIoU per round remains stable (66.6\%$\rightarrow$66.3\%), with reduced performance drop from easy to hard domains ($-5.3\%p$) in the $1^{st}$ round, indicating a stable adaptation throughout the experiment.



\subsection{Ablation Studies}


\begin{table}[]
    \scriptsize\addtolength{\tabcolsep}{0.3pt}
    \centering
    \caption{\textbf{Ablation Studies in Cityscapes-to-ACDC benchmark.} DDSD represents the experiment where DDSD participated in tuning method selection. Gain denotes the performance gain compared with (a). 
    }
    \vspace{-1em}
    \label{tab:ablation-acdc}
    \begin{tabular}{l|ccc|cc|c}
    \toprule
     & FT & DDSD & MIMA & mIoU & Gain & FPS \\
    \midrule
    (a) & \checkmark & & & 59.3$\pm$0.11 & & 1.976 \\
    (b) & \checkmark & & \checkmark & 61.9$\pm$0.147 & $+2.6\%p$ & 1.608 \\
    (c) & & \checkmark & & 59.8$\pm$0.19 & $+0.5\%p$ & 2.546 \\
    (d) - ours & & \checkmark & \checkmark & \textbf{62.2}$\pm$0.148 & $+2.9\%p$ & 2.004 \\
    \bottomrule
    \end{tabular}
    \vspace{-1.5em}
\end{table}




In Tab.~\ref{tab:ablation-acdc}, we conducted ablation studies to assess the impact of DDSD and MIMA on the Cityscapes-to-ACDC benchmark. Note that `FT' refers to experiments where only FT was used, and
the results are averaged over 5 runs with standard deviations being reported.
The results in (a) and (b) show that MIMA's self-supervised learning strategy significantly improves segmentation performance, yielding a $+2.6\%p$ gain. In (c), DDSD boosts segmentation performance  by $+0.5\%p$ and also increases frames per second (FPS) by 130\% (1.976$\rightarrow$2.546).
In (d), combining DDSD and MIMA results in a $+2.9\%p$ mIoU gain and a slight improvement in FPS (1.976$\rightarrow$2.004). This result aligns with the performance visualization in Fig.~\ref{fig:4.1-performance graph}, where (h) Hybrid-TTA (DDSD) mitigates the performance drop of (i) FT+MIMA during the last five rounds.
Further analysis on DDSD is discussed in Sec.~\ref{sec:ddsd-add-behaviour}.

\subsection{Analysis}
\noindent\textbf{Adaptation Time and Performance}
\label{sec:inference time}
Tab.~\ref{tab:fps} compares the performance and FPS of various methods on the Cityscapes-to-ACDC benchmark, which is also displayed in Fig.~\ref{fig:fps_comparison}.
10-round experiment, as in Tab.~\ref{tab:performance_comparison}, measured performance and average FPS. All experiments were carried out in a fair environment using NVIDIA RTX 6000 Ada.

(b) Hybrid-TTA achieves the best performance of 62.2\%, except for (a) Hybrid-TTA++, a variant of our method in which test-time augmentation~\cite{wang2022continual,yang2024exploring,liu2023vida} was applied. (a) shows the highest performance of 63.4\%, a $1.2\%p$ improvement due to enhanced pseudo-labeling via multiple augmentations.
(b) reaches 2.004 FPS while delivering outstanding performance (62.2\%).
In contrast, (h) TENT-continual shows poor performance due to its limited adaptation capacity. Despite being significantly faster, both (h) TENT-continual and (g) Source yield low results of 52.5\% and 56.9\%, respectively. (c) SVDP and (e) CoTTA achieve good segmentation performance (58.7\% and 61.0\%) but are impractical for online adaptation, requiring about 17 seconds and 10 seconds to process a single image. In (d) SVDP (w/o aug) and (f) CoTTA (w/o aug), where the test-time adaptation strategy is removed, FPS improves slightly at the cost of reduced performance, but their FPS remains significantly lower than ours.

\begin{table}[t!]
    \scriptsize\addtolength{\tabcolsep}{0.5pt}
    \centering
    \caption{{\bf Performance and FPS comparison on various methods.} 10-round experiment on Cityscapes-to-ACDC benchmark (as in Tab.~\ref{tab:performance_comparison}) is used to measure the performance and Frames Per Second (FPS). (a) Hybrid-TTA++ indicates that test-time augmentation strategy~\cite{wang2022continual} is additionally utilized for performance improvement, and (w/o aug) of (d), (f) means it is removed for FPS gain.}
    \vspace{-1em}
    \label{tab:fps}
    \begin{tabular}{lcccc}
    \toprule
    \textbf{Method} & \textbf{Tuning type} & \textbf{Performance (mIoU)} & \textbf{FPS} \\
    \midrule
    (a) Hybrid-TTA++                    & DDSD          & 63.4 & 0.136 \\
    (b) Hybrid-TTA (ours)               & DDSD          & 62.2 & 2.004 \\
    (c) SVDP~\cite{yang2024exploring}   & FT            & 61.0 & 0.058 \\
    (d) SVDP (w/o aug)                  & FT            & 59.8 & 0.076 \\
    (e) CoTTA~\cite{wang2022continual}  & FT            & 58.7 & 0.100 \\
    (f) CoTTA (w/o aug)                 & FT            & 57.6 & 0.555 \\
    (g) Source                          & no adaptation & 56.9 & 7.984 \\
    (h) TENT-continual~\cite{wang2020tent}& ET          & 52.5 & 3.609 \\
    \bottomrule
    \end{tabular}
    \vspace{-0.5cm} 
\end{table}

\begin{figure}[t!]
    \begin{subfigure}[b]{0.48\textwidth}
        \centering
        \includegraphics[width=\columnwidth]{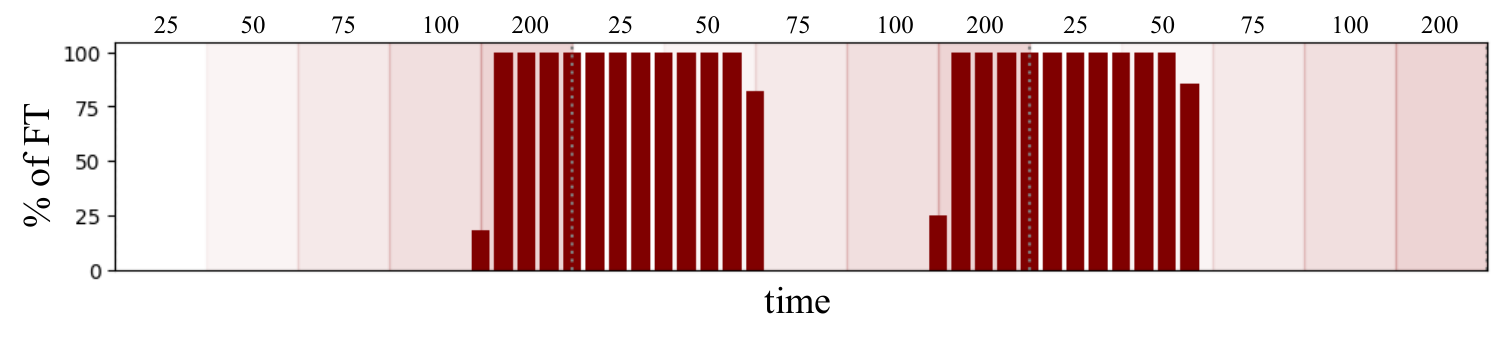}
        \vspace{-1.5em}
        \caption{Adaptive Domain Detection:   mIoU=64.7}
        \label{fig:compare-add}
    \end{subfigure}

    \begin{subfigure}[b]{0.48\textwidth}
        \centering
        \includegraphics[width=\columnwidth]{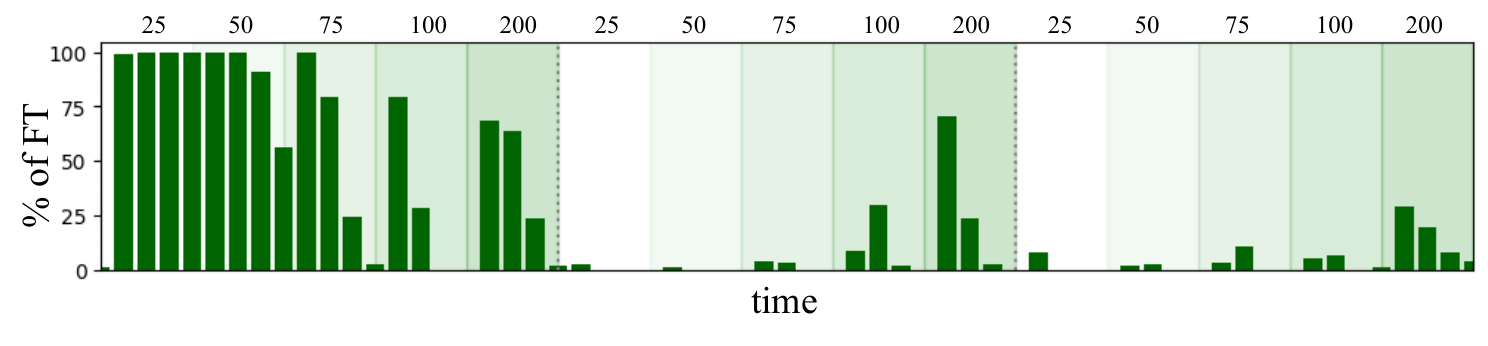}
        \vspace{-1.5em}
        \caption{DDSD:   mIoU=66.73}
        \label{fig:compare-ddsd}
    \end{subfigure}
    \vspace{-0.7cm}
    \caption{{\bf Performance comparison of Domain Shift Detection methods.} (a) is Adaptive Domain Detection~\cite{colomer2023adapt} and (b) is ours.} 
    \label{fig:compare-dsd}
    \vspace{-0.6cm}
\end{figure}

\noindent\textbf{Domain Shift Detection}
\label{sec:ddsd-add-behaviour}
Fig.~\ref{fig:compare-dsd} compares the behaviors of two domain shift detectors, DDSD and Adaptive Domain Detection (ADD) proposed in HAMLET~\cite{colomer2023adapt}, to examine their initial adaptation to new domains and reaction afterwards. We conducted cyclic adaptation using 5 rain intensities (25mm-200mm) 3 times on the OnDA benchmark.
This benchmark, with more subtle domain shifts, is used to evaluate the sensitivity of domain shift detectors.
The changing background colors indicate domain changes, while the y-axis shows the ratio of FT activation every 125 timesteps, \emph{e.g.}, if FT is activated 100 times out of 125 timesteps and AT 25 times, the ratio is 80\%.

{Fig.~\ref{fig:compare-add} illustrates the behavior of ADD, which detects domain shifts by comparing the feature distance between the student model and ImageNet pretrained model. When a domain shift is detected, ADD activates FT for a fixed time window (\emph{e.g.}, 1000 timesteps) for adaptation. We observe that ADD sticks to AT for the first 4 domains in the first round without detecting any shifts, and then detects the first domain shift at the start of 200mm domain, triggering FT for a certain timesteps. This is because ADD uses a fixed threshold to detect domain shifts, ignoring shifts with gaps less than 200mm. This static thresholding limits the the model's adaptability, leading to suboptimal performance with an mIoU of 64.7\%.
Moreover, we noticed that the FT was being activated longer than expected, until 25mm and 50mm domains in the second round. This overflow happens due to the fixed FT activation window of ADD~\cite{colomer2023adapt}, highlighting the need for instance-wise domain shift detection.}

In contrast, Fig.~\ref{fig:compare-ddsd} shows that DDSD activates FT at the beginning of the experiment, as it assumes that the model faces a large domain shift between source and target domains.
After the first round, DDSD begins to prefer AT over FT, resulting in a reduced FT ratio.
In the 200mm domain of the $2^{nd}$ round, the FT ratio dramatically increases, as 200mm is farther from source domain requiring additional adaptation. This happens again in the third round, but with significantly reduced FT ratio. This indicates that DDSD evolves over time with the changing target domains, leading to a superior mIoU of 66.7\%.

\section{Conclusion}

In this paper, we have proposed Hybrid-TTA that dynamically selects the optimal tuning method for each input instance. This framework promotes the interaction between the model and ever-changing input sequences on target domain, resulting in improved performance. By integrating an auxiliary reconstruction task within segmentation CTTA, our model extracts generalized features and maintains robustness across domains. Experiments demonstrate that our method achieves outstanding performance in semantic segmentation with significantly reduced computational overhead. It is expected that our hybrid framework provides new insights on developing reliable techniques for CTTA.

{
    \small
    \bibliographystyle{ieeenat_fullname}
    \bibliography{main}
}

\clearpage
\setcounter{section}{0}
\maketitlesupplementary

\definecolor{ronecolor}{RGB}{255, 69, 58}
\definecolor{rthreecolor}{RGB}{58, 134, 255}
\definecolor{rfourcolor}{RGB}{34, 139, 34}

\newcommand{\Rone}{\textcolor{ronecolor}{\textbf{R1}}\xspace}
\newcommand{\Rtwo}{\textcolor{rthreecolor}{\textbf{R2}}\xspace}
\newcommand{\Rthree}{\textcolor{rfourcolor}{\textbf{R3}}\xspace}

\section{Implementation Details}

We report the implementation details used to train in the described method, including network architectures, and hyperparameters.


\paragraph{Model Architecture}
For semantic segmentation, we use SegFormer MiT-B5 for feature extractor $f_\theta$ and segmentation decoder $h_\phi$.
For reconstruction, SimMIM~\cite{xie2022simmim} is used for the implementation of reconstruction decoder $g_\psi$.
To implement Adapter Tuning, we utilize AdaptMLP from \cite{chen2022adaptformer} into each transformer layers to implement efficient tuning, where Full Tuning updates all parameters including the adapter, and Adapter Tuning updates only the adapter parameters (which occupies 0.1\% of the entire parameters).


\paragraph{Hyperparameters}
For test-time adaptation, we set the batch size to 1, and used Adam optimizer with learning rate of $6\times10^{-5}/8$, in reference of \cite{wang2022continual,liu2023vida}. For DDSD implementation, we set the initial DDSD threshold $\tau$ to be 0, and default $\alpha_l$ to be 0.999, following \cite{tarvainen2017mean}. For MIMA implementation, we masked the images with masking ratio 0.6 and mask patch size 32, as in \cite{xie2022simmim}. Unlike recent CTTA studies \cite{wang2022continual,liu2023vida,yang2024exploring} that used multi-scale input with flipping as the test-time Augmentation, we did not use any augmentation strategy for our main results.
$\lambda_d$=1.1 and $\lambda_r$=0.3 were heuristically selected based on a small subset of the first target domain and kept fixed across all experiments to avoid test-time overfitting. The same hyperparameters are used for both benchmarks.
\section{Ablation Study}

\paragraph{Hyperparameter $\lambda_{d}$}

\begin{figure}[ht]
    \begin{subfigure}[b]{0.48\textwidth}
        \centering
        \includegraphics[width=\columnwidth]{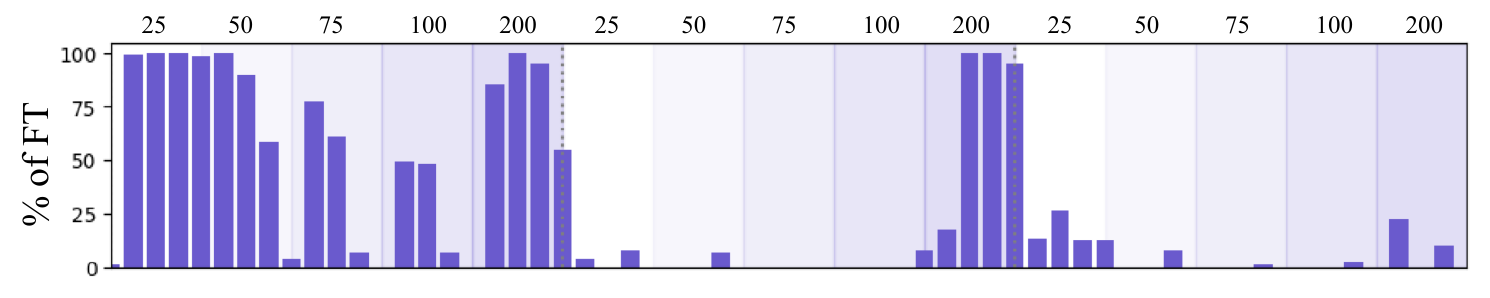}
        \caption{$\lambda_{d}$=1.1}
        \label{fig:lambda11}
    \end{subfigure}
    \begin{subfigure}[b]{0.48\textwidth}
        \centering
        \includegraphics[width=\columnwidth]{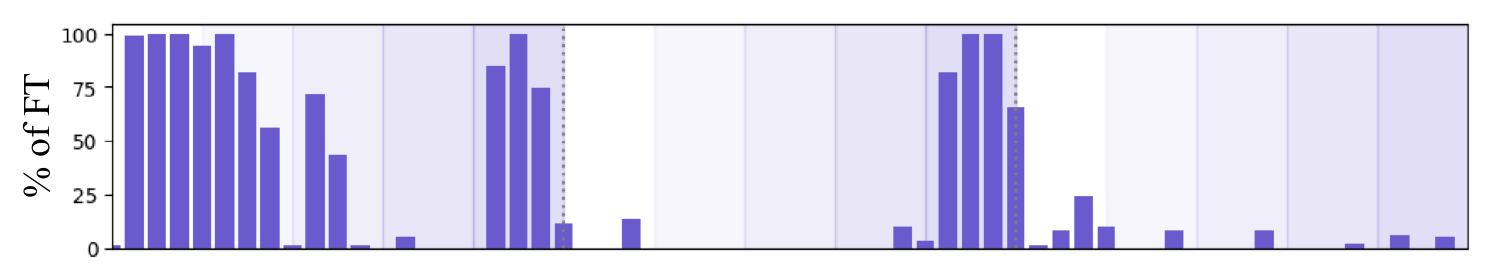}
        \caption{$\lambda_{d}$=1.2}
        \label{fig:lambda12}
    \end{subfigure}
    \begin{subfigure}[b]{0.48\textwidth}
        \centering
        \includegraphics[width=\columnwidth]{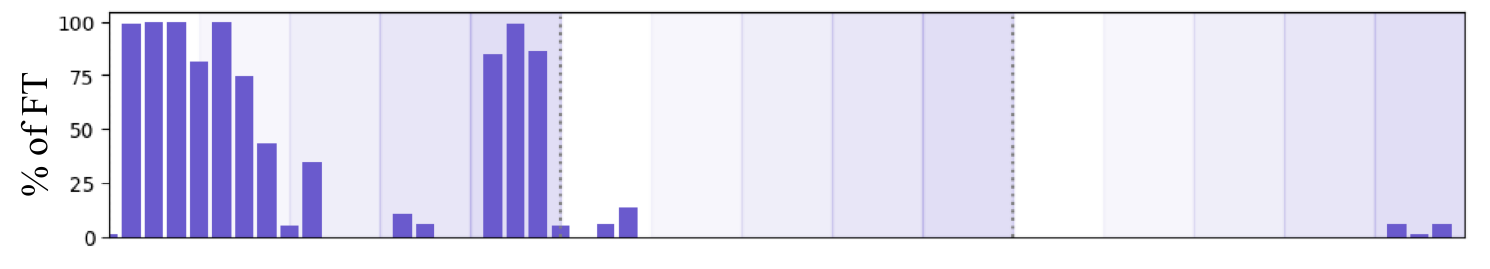}
        \caption{$\lambda_{d}$=1.3}
        \label{fig:lambda13}
    \end{subfigure}
\caption{{\bf Ablation on hyperparameter $\lambda_{d}$.} Percentage of FT usage in OnDA benchmark. x-axis represents time and y-axis represents FT usage ratio. 5 domains are cyclically repeated 3 times.}
\label{fig:lambda_ablation}
\end{figure}

We conducted ablation experiments to investigate the role the hyperparameter $\lambda_{d}$ in DDSD works within the system, using OnDA benchmark~\cite{panagiotakopoulos2022online}. Only the segmentation loss $L_{seg}$ has engaged in this ablation study to exclude the influence of MIMA. We performed cyclic adaptation over 5 rain intensities (25mm-200mm) 3 times, to understand how DDSD adapts to a completely new domains and how it reacts after some initial adaptations. Since 25mm is closer to the source domain and 200mm farther, we consider 200mm to be harder to adapt to, requiring more adaptation than 25mm. The domain changes are visualized by changing background colors, according to the changing rain intensities written at the top. The height of each bar represents the ratio of Full Tuning (FT) being activated in every 125 timesteps, \emph{e.g.,} if FT is activated 100 times out of 125, the ratio is 80\%.

Fig~\ref{fig:lambda_ablation} illustrates how DDSD evolves over time. After aggressive adaptation in the first round, triggered by the initial encounter with the target domain, DDSD begins to favour AT over FT, leading to a reduced FT ratio. This change occurs because the target domain becomes relatively familiar to both the student and teacher models.
However, in 200mm domain during the first round, the FT ratio increases, as 200mm is significantly more distant from any other domains and thus exhibits lower temporal correlation. This phenomenon consistently holds across all experiments, regardless of the size of $\lambda_d$ value.

The effect of hyperparameter $\lambda_d$ is particularly evident in 100mm domain during the first round. In Fig.~\ref{fig:lambda11}, the FT ratio is nearly 50\%, while in Fig.~\ref{fig:lambda12} and Fig.~\ref{fig:lambda13}, the ratio drops sharply to below 10\%. Similarly, in 200mm domain during the second round, FT is not activated in Fig.~\ref{fig:lambda13}, unlike the other two graphs.

From these observations, we can conclude that as $\lambda_d$ increases, DDSD becomes less sensitive to domain shifts. This is due to the role of $\lambda_d$ a sensitivity hyperparameter (described in Sec.~\ref{sec:ddsd}), which acts as a scaling factor for the dynamic threshold $\tau_t$. As the dynamic threshold increases, a larger loss is required to activate FT, and vice versa.




\begin{table*}[hbt!]
\scriptsize\addtolength{\tabcolsep}{-2.0pt}
\centering
\caption{Ablation Study in Cityscapes-to-ACDC benchmark for \textbf{20 rounds.} Mean is the average score of mIoU.}
\label{tab:cityscapes-to-acdc-20rounds}
\begin{tabular}{l|ccc|cccc|cccc|cccc|ll}
\toprule
\bf{Test} & & &
    & \multicolumn{4}{c}{1}
    & \multicolumn{4}{c}{10}
    & \multicolumn{4}{c|}{20}
    & \multicolumn{2}{c}{Mean $\uparrow$} \\
\bf{Condition} & FT & DDSD & MIMA
    & \textbf{F} & \textbf{N} & \textbf{R} & \textbf{S}
    & \textbf{F} & \textbf{N} & \textbf{R} & \textbf{S}
    & \textbf{F} & \textbf{N} & \textbf{R} & \textbf{S}
    & 10R & 20R \\
\midrule
(a) & \checkmark & &          
    & 70.6 & 42.7 & 62.7 & 61.7
    & 69.3 & 42.7 & 61.2 & 59.7
    & 65.5 & 39.3 & 55.9 & 58.5
    & 59.3 & 55.9 \\
(b) & \checkmark & & \checkmark 
    & 70.3 & 44.5 & 65.1 & 63.3
    & 68.9 & 49.2 & 64.9 & 61.5
    & 66.2 & 48.2 & 62.0 & 57.5
    & 61.9 & 59.6 \\
(c) & & \checkmark & 
    & 70.9 & 43.2 & 64.0 & 60.7
    & 70.5 & 43.0 & 64.3 & 60.5
    & 69.6 & 43.6 & 62.5 & 59.1
    & 59.8 & 58.9 \\
(d) - ours & & \checkmark & \checkmark 
    & 70.3 & 44.5 & 65.1 & 63.2
    & 69.9 & 49.5 & 66.5 & 63.0
    & 68.8 & 50.1 & 63.9 & 59.9
    & 62.2 & 61.5 \\
\bottomrule
\end{tabular}
\end{table*}



\paragraph{Hyperparameter $\lambda_r$}

\begin{table}[t!]
    \scriptsize\addtolength{\tabcolsep}{0.5pt}
    \centering
    \caption{{\bf Ablation on hyperparameter $\lambda_r$} Performance in Cityscapes-to-ACDC benchmark.}
    \label{tab:ablation-lambda_r}
    \begin{tabular}{ccc}
    \toprule
    & \textbf{$\lambda_r$} & \textbf{Performance (mIoU)} \\
    \midrule
    (a)             & 0.0           & 61.3 \\
    \textbf{(b)}    & \textbf{0.3}  & \textbf{61.9} \\
    (c)             & 0.5           & 61.3 \\
    (d)             & 1.0           & 60.6 \\
    \bottomrule
    \end{tabular}
\end{table}

In Tab.~\ref{tab:ablation-lambda_r}, we conducted ablation studies with respect to $\lambda_r$ in \eqref{eqn:3.3.2}, which determines the influence of the reconstruction loss $\mathcal{L}^{rec}_{\theta,\psi}$. To exclude the influence of DDSD, we only utilized Full Tuning.

In (a), $\lambda_r$ is set to 0, meaning that the reconstruction loss does not participate in test-time Adaptation process. This setup is similar to MIC~\cite{hoyer2023mic}, where the model is adapted solely through the consistency loss between student and teacher prediction maps. Since some important visual information is masked out from the target image, (a) yields a suboptimal performance of 61.3\%.
In (b), $\lambda_r$ is set to 0.3, achieving the best performance.
However, as $\lambda_r$ increases in (c) and (d), performance declines. This degradation occurs because the influence of the segmentation loss diminishes as reconstruction loss dominates the adaptation process with big $\lambda_r$. Consequently, adaptation towards segmentation, which is our primary task, becomes weaker.


\paragraph{Long-term Adaptation}

Although 0.5\%p improvement of DDSD in Tab.~\ref{tab:ablation-acdc} may appear minor, we believe its core strength--stability--has been underestimated within the scope of the original 10-round experiment.
Fig.~\ref{fig:4.1-performance graph} clearly illustrates that DDSD maintains stable performance, whereas Full Tuning (FT) exhibits a sharp performance decline after the 5\textsuperscript{th} round, indicating that DDSD’s advantage will grow over time.
To further support this observation, we extend the ablation study reported in Tab.~\ref{tab:ablation-acdc} from 10 rounds to 20 rounds. As shown in Tab.~\ref{tab:cityscapes-to-acdc-20rounds}, (c) DDSD consistently outperforms FT at both the 10\textsuperscript{th} and 20\textsuperscript{th} rounds, with an widening performance gap (0.5$\rightarrow$\textbf{3.0} between (a) FT and (c) DDSD, 0.3$\rightarrow$\textbf{1.9} between (b) FT+MIMA and (d) DDSD+MIMA). These results strongly confirm the superior long-term stability of DDSD.



\begin{table}
    \scriptsize\addtolength{\tabcolsep}{-2pt}
    \centering
    \caption{Performance comparison of variants of proposed method on \textbf{Source Dataset} (CityScapes). \emph{ours} refers to Hybrid-TTA, our main result, \emph{Source} refers to SegFormer MiT-B5 with no adaptation.}
    \label{tab:source_forgetting_supp}
    \begin{tabular}{l|ccccc|cc}
        \toprule
            \bf{Test} & \textbf{1} & \textbf{2} & \textbf{3} & \textbf{4} & \textbf{5} & \textbf{Mean} & \textbf{Target Mean} \\
        \midrule
        (a) Source
            & 78.0 & 78.0 & 78.0 & 78.0 & 78.0 & 78.0 & 62.4 \\
        (b) AT          
            & 77.7 & 77.4 & 77.2 & 77.3 & 77.0 & \textbf{77.3} & 60.5 \\
        (c) DDSD        
            & 76.7 & 76.2 & 75.6 & 75.4 & 75.2 & 75.8 & 64.2 \\
        \textbf{(d) DDSD+MIMA} 
            & 77.0 & 76.0 & 75.3 & 75.0 & 74.7 & 75.5 & \textbf{66.7} \\
        (e) FT          
            & 76.5 & 75.6 & 75.2 & 74.9 & 74.7 & 75.4 & 65.0 \\
        (f) AT+MIMA     
            & 77.3 & 75.9 & 74.8 & 74.0 & 73.1 & 75.0 & 65.3 \\
        (g) FT+MIMA     
            & 76.8 & 75.7 & 74.8 & 74.2 & 73.7 & 75.0 & 66.3 \\
        \bottomrule
    \end{tabular}
\end{table}


\section{Analysis}



\paragraph{Source Domain Forgetting}

Tab.~\ref{tab:source_forgetting_supp} presents the performance on Cityscapes dataset after each round of adaptation on OnDA benchmark. Since Cityscapes serves as the source dataset for adaptation, we aim to assess the severity of catastrophic forgetting after each round of adaptation. `Mean' represents the mIoU performance on Cityscapes, while `Target Mean' refers to the performance on synthetic rain dataset, provided for comparison.

(a) Source, SegFormer MiT-B5 without any adaptation, serves as the upper bound, achieving 78\% on the source dataset, with moderate target performance of 62.4\%.
(b) AT, which updates only the adapter parameters, achieves the best source performance of 77.3\%, but its target performance (60.5\%) is lower than Source. This is because AT effectively preserves source knowledge and prevents catastrophic forgetting but has limited adaptability.
However, this limitation is significantly mitigated by incorporating MIMA into AT, as (f) AT+MIMA shows 65.3\% of target performance at the cost of slight reduction in source performance.

Here, we can assume that MIMA encourages the model to lose the source knowledge and collect target knowledge, as a similar pattern is observed with (e) FT and (g) FT+MIMA, where (g) exhibits lower source performance but higher target performance compared to FT. As seen from these findings, losing source knowledge is not necessarily disadvantageous, because the model learns as much as it forgets.

Nevertheless, excessive loss of source knowledge can degrade target performance. Notably, (d) DDSD+MIMA (ours) achieves an impressive 66.7\% mIoU on the target dataset, with a gain of $+4.3\%p$ compared to Source, while substantially preserving source performance ($-2.5\%p$).


\begin{table*}[hbt!]
\caption{Performance comparison in \textbf{OASIS benchmark}. GTA~\cite{richter2016playing} as the source dataset, and ACDC~\cite{sakaridis2021acdc} as the test dataset.}
\label{tab:oasis-gta}
\scriptsize\addtolength{\tabcolsep}{-2.0pt}
\centering
\begin{tabular}{l|cccc|cccc|cccc|c}
\toprule
\bf{Test}
    & \multicolumn{4}{c}{1}
    & \multicolumn{4}{c}{2}
    & \multicolumn{4}{c|}{3}
    & \\
\bf{Condition}
    & \textbf{F} & \textbf{N} & \textbf{R} & \textbf{S}
    & \textbf{F} & \textbf{N} & \textbf{R} & \textbf{S}
    & \textbf{F} & \textbf{N} & \textbf{R} & \textbf{S}
    & Mean $\uparrow$\\
\midrule
Source 
    & 43.4 & 19.6 & 41.3 & 38.1
    & 43.4 & 19.6 & 41.3 & 38.1
    & 43.4 & 19.6 & 41.3 & 38.1
    & 35.6 \\
TENT~\cite{wang2020tent} 
    & 43.8 & 19.8 & 42.1 & 38.7
    & 43.9 & 18.8 & 39.9 & 35.9
    & 43.0 & 17.1 & 37.4 & 33.5
    & 34.5 \\
SVDP~\cite{yang2024exploring} 
    & \textbf{46.9} & \textbf{25.2} & 45.6 & 41.8
    & 48.0 & 25.0 & 43.8 & 39.6
    & 45.1 & 22.5 & 42.6 & 40.1
    & 38.8 \\
C-MAE~\cite{liu2024continual} 
    & 46.1 & 20.0 & 43.2 & 38.9
    & 45.8 & 18.9 & 42.9 & 39.1
    & 46.2 & 20.5 & 43.2 & 38.1
    & 36.9 \\
\midrule
\textbf{Ours} 
    & 45.1 & 23.0 & \textbf{46.6} & \textbf{42.5}
    & \textbf{48.9} & \textbf{26.9} & \textbf{48.2} & \textbf{43.3}
    & \textbf{49.2} & \textbf{28.8} & \textbf{48.0} & \textbf{43.3}
    & \textbf{41.2} \\
\bottomrule
\end{tabular}
\end{table*}


\begin{table*}[hbt!]
\scriptsize\addtolength{\tabcolsep}{-2.0pt}
\centering
\caption{Performance Comparison on \textbf{Cityscapes-to-ACDC benchmark} over 3 rounds.}
\label{tab:cityscapes-to-acdc-cmae}
\begin{tabular}{l|cccc|cccc|cccc|c}
\toprule
\bf{Test}
    & \multicolumn{4}{c}{1}
    & \multicolumn{4}{c}{2}
    & \multicolumn{4}{c|}{3}
    & \\
\bf{Condition}
    & \textbf{F} & \textbf{N} & \textbf{R} & \textbf{S}
    & \textbf{F} & \textbf{N} & \textbf{R} & \textbf{S}
    & \textbf{F} & \textbf{N} & \textbf{R} & \textbf{S}
    & Mean$\uparrow$ \\
\midrule
CoTTA \cite{wang2022continual}
    & 70.9 & 41.2 & 62.4 & 59.7
    & 70.9 & 41.1 & 62.6 & 59.7
    & 70.9 & 41.0 & 62.7 & 59.7
    & 58.6 \\
VDP \cite{gan2023decorate}
    & 70.5 & 41.1 & 62.1 & 59.5
    & 70.4 & 41.1 & 62.2 & 59.4
    & 70.4 & 41.0 & 62.2 & 59.4
    & 58.2 \\
BeCoTTA~\cite{lee2024becotta} 
    & \textbf{72.3} & 42.0 & 63.5 & 60.1
    & \textbf{72.4} & 41.9 & 63.5 & 60.2
    & \textbf{72.3} & 41.9 & 63.6 & 60.2
    & 59.5 \\
C-MAE~\cite{liu2024continual}
    & 71.9 & \textbf{44.6} & \textbf{67.4} & \textbf{63.2}
    & 71.7 & 44.9 & 66.5 & 63.1
    & \textbf{72.3} & 45.4 & \textbf{67.1} & 63.1
    & 61.8 \\
\midrule
\textbf{Ours}
    & 70.3 & 44.5 & 65.1 & \textbf{63.2}
    & 71.8 & \textbf{48.2} & \textbf{67.1} & \textbf{63.7}
    & 71.2 & \textbf{49.3} & \textbf{67.1} & \textbf{63.3}
    & \textbf{62.2} \\
\bottomrule
\end{tabular}
\end{table*}

\paragraph{GTA-to-ACDC benchmark}
We now dive deeper in Hybrid-TTA performance on the OASIS~\cite{volpi2022road} benchmark protocol, as detailed in Tab.~\ref{tab:oasis-gta}. The OASIS protocol involves training models on a synthetic source dataset (GTA~\cite{richter2016playing}), tuning hyperparameters on a synthetic validation dataset (SYNTHIA~\cite{ros2016synthia}), and finally evaluating model performance on a real-world test dataset (ACDC~\cite{sakaridis2021acdc}).
It is widely acknowledged that the OASIS benchmark presents greater challenge than the Cityscapes-to-ACDC benchmark due to a significantly larger domain gap between source and target datasets.

Our base model was trained on GTA following the training protocol described in \cite{volpi2022road}.
We select the best hyperparameters in SYNTHIA dataset as follows: $\lambda_d$=1.2, $\lambda_r$=0.2.
Finally, we present CTTA results on ACDC dataset over 3 rounds, where Ours outperforms SoTA~\cite{liu2024continual,yang2024exploring}, demonstrating our robustness under various environments.

Although using FT under large domain shifts may seem counter-intuitive,
severe shifts require greater adaptability than what efficient tuning can offer (\emph{e.g.}, TENT in Tab.~\ref{tab:oasis-gta}), at the cost of stability and with the forgetting risks associated with FT.
To manage this trade-off and achieve balance,
DDSD selectively triggers FT under significant distribution shifts, while avoiding unnecessary updates in stable regimes.
\paragraph{Performance Comparison with Parameter-efficient Fine-tuning Methods}

Tab.~\ref{tab:cityscapes-to-acdc-cmae} provides a detailed comparative study of various CTTA strategies on the Cityscapes-to-ACDC benchmark, extending the results discussed in \ref{sec:main_results}, but evaluated over 3 adaptation rounds to highlight early-stage adaptation behaviors and performances.

VDP~\cite{gan2023decorate}, a pioneering work that introduces a prompt-based Parameter-efficient Fine-tuning (PEFT) strategy for CTTA, achieves a relatively disappointing result of 58.2, notably underperforming even the baseline CoTTA. This suggests that prompt-based PEFT adaptation strategies might struggle in scenarios involving significant domain shifts, such as from CityScapes to ACDC.

BeCoTTA~\cite{lee2024becotta}, which employs a Mixture-of-Experts (MoE) based Parameter-efficient Fine-Tuning CTTA strategy, demonstrates slightly better average mIoU (59.5) compared to VDP (59.4).


\paragraph{Performance Comparison with Continual-MAE}

Tab.~\ref{tab:cityscapes-to-acdc-cmae} also provides performance of Continual-MAE~\cite{liu2024continual}, another pioneering strategy based on MIM, which attains a considerably stronger performance of 61.8.
While our method employing MIMA outperforms C-MAE, our method also differs significantly in methodology. Specifically, C-MAE reconstructs HOG features to emphasize geometric changes (\emph{e.g.}, shape), promoting the learning of domain-invariant features. On the other hand, MIMA reconstructs RGB values, not only enhancing robust feature extraction, but also capturing domain-specific low-level cues (\emph{e.g.}, color, texture, brightness). This enables the model to better capture cross-domain feature discrepancies, allowing DDSD to detect domain shifts more effectively.




While MIM has been previously explored in TTA as a domain-agnostic regularizer~\cite{liu2024continual} or pseudo-label generator~\cite{mansour2024ttt}, our work departs from these conventional usages in both design and intent. The integration between MIMA and DDSD is not merely synergistic but functionally complementary: MIMA enhances sensitivity to domain discrepancies while improving robustness on familiar data, and DDSD leverages this sensitivity to selectively detect domain shifts.
We believe this task-driven coupling of MIM and domain shift detection in CTTA is novel, offering a new perspective on how reconstruction-based signals can be actively utilized beyond pretraining.

\section{Performance Comparison (Full scores)}

We provide the entire performance results of segmentation CTTA experiments in Tab.~\ref{tab:performance_comparison_supp} and Tab.~\ref{tab:performance_comparison_supp_onda}.
Our proposed method, Hybrid-TTA, achieves 0.6\%p mIoU improvement over the previous state-of-the-art method on Cityscapes-to-ACDC benchmark. Moreover, it outperforms other CTTA methods on OnDA benchmark by 0.9\%p. Notably, Hybrid-TTA also achieves more than 20 times higher FPS than any other CTTA method with comparable mIoU performance, including CoTTA and SVDP (See Fig.~\ref{fig:fps_comparison} and Tab.~\ref{tab:fps}), offering a robust solution for real-world online continual adaptation challenges.


\section{Qualitative Results}

\begin{figure*}[t!]
\vspace{7em}
\centering
\includegraphics[width=2.1\columnwidth]{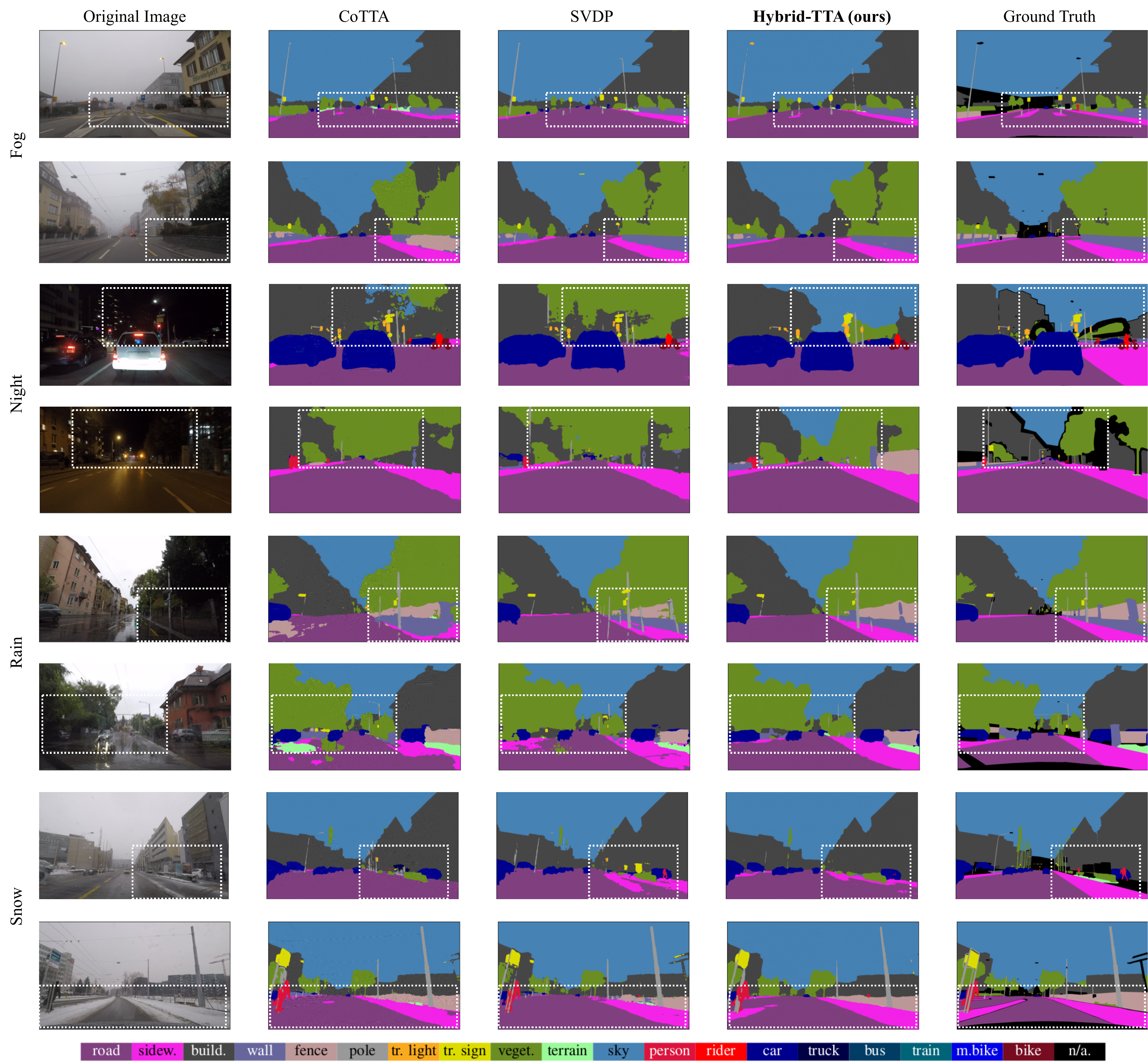}
\caption{\textbf{Qualitative comparison of segmentation results on Fog, Night, Rain and Snow domains for Cityscapes-to-ACDC benchmark.}}
\label{fig:qualitative}
\vspace{5em}
\end{figure*}

Fig.~\ref{fig:qualitative} is qualitative comparison of segmentation results on Fog, Night, Rain and Snow domains for Cityscapes-to-ACDC benchmark. Hybrid-TTA (column 4) is showing remarkable segmentation results compared to other methods including CoTTA \cite{wang2022continual} and SVDP \cite{yang2024exploring}, notably SVDP being currently the state-of-the-art method in semantic segmentation CTTA.

In Night domain (row 3-4), Hybrid-TTA shows outstanding performance as it is demonstrated in the main paper. It is noticeable in Hybrid-TTA (row 3-4, column 4), particularly in distinguishing the sky from vegetation (green) and buildings (gray), where lighting conditions are poor. This is a significant achievement, given that other methods often misclassify these elements due to the altered appearance of objects at night.

For Rain domain (row 5-6), Hybrid-TTA excels in segmenting fine details such as fence (beige) and accurately identifying cars (deep blue) and sidewalk (pink) from road (purple), which other methods often confuse with vegetation (green) or terrain (light green). This highlights the model's ability to maintain clear boundaries and accurate object classification under adverse weather conditions.

In case of Snow domain (row 7-8), Hybrid-TTA effectively delineates sidewalk (pink) and other urban features, producing sharper segmentation maps compared to CoTTA and SVDP, which often blur these boundaries.

Overall, we can observe that Hybrid-TTA not only maintains robustness across diverse environmental conditions but also mitigate common segmentation issues, such as dirty segmentation maps observed in CoTTA (row 3, column 2) and SVDP (row 6, column 3). This robustness is particularly evident in the Night domain, which is typically considered the most challenging due to its low contrast and ambiguous object boundaries. These qualitative results underscore the effectiveness of Hybrid-TTA in real-world scenarios, where adaptability and precision are crucial.


\begin{landscape}
\begin{table}[p]
    \scriptsize\addtolength{\tabcolsep}{-1.2pt}
    \centering
    \caption{Performance comparison of Segmentation CTTA baselines on \textbf{Cityscapes-to-ACDC benchmark} over 10 cyclic rounds. Mean is the average score of mIoU.}
    \label{tab:performance_comparison_supp}
    \begin{tabular}{l|llll|llll|llll|llll|llll|l}
        \toprule
        \bf{Test} &
            \multicolumn{4}{c}{1} &
            \multicolumn{4}{c}{2} &
            \multicolumn{4}{c}{3} &
            \multicolumn{4}{c}{4} &
            \multicolumn{4}{c|}{5} & \\
        \bf{Condition} &
            \textbf{Fog} & \textbf{Night} & \textbf{Rain} & \textbf{Snow} &
            \textbf{Fog} & \textbf{Night} & \textbf{Rain} & \textbf{Snow} &
            \textbf{Fog} & \textbf{Night} & \textbf{Rain} & \textbf{Snow} &
            \textbf{Fog} & \textbf{Night} & \textbf{Rain} & \textbf{Snow} &
            \textbf{Fog} & \textbf{Night} & \textbf{Rain} & \textbf{Snow} &
            Cont. \\
        \midrule
        Source
            & 69.1 & 40.3 & 59.7 & 57.8
            & 69.1 & 40.3 & 59.7 & 57.8
            & 69.1 & 40.3 & 59.7 & 57.8
            & 69.1 & 40.3 & 59.7 & 57.8
            & 69.1 & 40.3 & 59.7 & 57.8
            & - \\
        CoTTA \cite{wang2022continual}
            & 70.9 & 41.2 & 62.4 & 59.7
            & 70.9 & 41.1 & 62.6 & 59.7
            & 70.9 & 41.0 & 62.7 & 59.7
            & 70.9 & 41.0 & 62.7 & 59.7
            & 70.9 & 41.0 & 62.8 & 59.7
            & - \\
        SVDP \cite{yang2024exploring}
            & 71.0 & 43.2 & 64.8 & 62.0
            & 71.9 & 44.3 & 66.1 & 62.5
            & 72.2 & 44.1 & 66.6 & 62.5
            & 71.8 & 43.9 & 66.6 & 62.3
            & 72.1 & 43.7 & 66.7 & 62.5
            & - \\
        ViDA \cite{liu2023vida}
            & 71.6 & 43.2 & 66.0 & 63.4
            & 73.2 & 44.5 & 67.0 & 63.9
            & 73.2 & 44.6 & 67.2 & 64.2
            & 70.9 & 44.0 & 66.0 & 63.2
            & 72.0 & 43.7 & 66.3 & 63.1
            & - \\
        \midrule
        \bf{Hybrid-TTA}
            & 70.3 & 44.5 & 65.1 & 63.2
            & 71.8 & 48.2 & 67.1 & 63.7
            & 71.2 & 49.3 & 67.1 & 63.3
            & 70.1 & 49.3 & 66.9 & 63.1
            & 69.5 & 49.2 & 66.6 & 63.0
            & - \\
        \bottomrule
    \end{tabular}

    \begin{tabular}{l|llll|llll|llll|llll|llll|l}
        \toprule
        \bf{Test} &
            \multicolumn{4}{c}{6} &
            \multicolumn{4}{c}{7} &
            \multicolumn{4}{c}{8} &
            \multicolumn{4}{c}{9} &
            \multicolumn{4}{c|}{10} &
            All$\uparrow$ \\
        \bf{Condition} & 
            \textbf{Fog} & \textbf{Night} & \textbf{Rain} & \textbf{Snow} &
            \textbf{Fog} & \textbf{Night} & \textbf{Rain} & \textbf{Snow} &
            \textbf{Fog} & \textbf{Night} & \textbf{Rain} & \textbf{Snow} &
            \textbf{Fog} & \textbf{Night} & \textbf{Rain} & \textbf{Snow} &
            \textbf{Fog} & \textbf{Night} & \textbf{Rain} & \textbf{Snow} &
            Mean \\
        \midrule
        Source
            & 69.1 & 40.3 & 59.7 & 57.8
            & 69.1 & 40.3 & 59.7 & 57.8
            & 69.1 & 40.3 & 59.7 & 57.8
            & 69.1 & 40.3 & 59.7 & 57.8
            & 69.1 & 40.3 & 59.7 & 57.8
            & 56.9 \\
        CoTTA \cite{wang2022continual}
            & 70.9 & 41.0 & 62.8 & 59.7
            & 70.9 & 41.1 & 62.6 & 59.7
            & 70.9 & 41.1 & 62.6 & 59.7
            & 70.8 & 41.1 & 62.6 & 59.7
            & 70.8 & 41.1 & 62.6 & 59.7
            & 58.6 \\
        SVDP \cite{yang2024exploring}
            & 71.9 & 43.9 & 66.8 & 62.4
            & 71.9 & 43.6 & 66.5 & 62.3
            & 71.6 & 43.3 & 66.6 & 62.1
            & 71.7 & 43.7 & 66.5 & 61.5
            & 71.6 & 43.5 & 66.6 & 62.0
            & 61.0 \\
        ViDA \cite{liu2023vida}
            & 72.2 & 44.0 & 66.6 & 62.9
            & 72.3 & 44.8 & 66.5 & 62.9
            & 72.1 & 45.1 & 66.2 & 62.9
            & 71.9 & 45.3 & 66.3 & 62.9
            & 72.2 & 45.2 & 66.5 & 62.9
            & 61.6 \\
        \midrule
        \bf{Hybrid-TTA}
            & 69.5 & 49.3 & 66.7 & 63.0
            & 69.6 & 49.4 & 66.7 & 63.0
            & 69.7 & 49.5 & 66.7 & 63.0
            & 69.7 & 49.4 & 66.6 & 63.1
            & 69.9 & 49.5 & 66.5 & 63.0
            & \textbf{62.2} \\
        \bottomrule
    \end{tabular}
\end{table}

\begin{table}[bp]
    \scriptsize\addtolength{\tabcolsep}{-1.2pt}
    \centering
    \caption{Performance comparison of Segmentation CTTA baselines on \textbf{OnDA benchmark.} over 5 cyclic rounds. Mean is the average score of mIoU.}
    \label{tab:performance_comparison_supp_onda}

    \begin{tabular}{l|ccccc|ccccc|ccccc|ccccc|ccccc|l}
        \toprule
            \bf{Test} & \multicolumn{5}{c|}{1} & \multicolumn{5}{c}{2|} & \multicolumn{5}{c|}{3} & \multicolumn{5}{c|}{4} & \multicolumn{5}{c|}{5} & All$\uparrow$ \\
            \bf{Intensity (mm)}
                & \textbf{25} & \textbf{50} & \textbf{75} & \textbf{100} & \textbf{200}
                & \textbf{25} & \textbf{50} & \textbf{75} & \textbf{100} & \textbf{200}
                & \textbf{25} & \textbf{50} & \textbf{75} & \textbf{100} & \textbf{200}
                & \textbf{25} & \textbf{50} & \textbf{75} & \textbf{100} & \textbf{200}
                & \textbf{25} & \textbf{50} & \textbf{75} & \textbf{100} & \textbf{200}
                & Mean \\
        \midrule
        Source
            & 67.7 & 65.6 & 62.9 & 58.1 & 45.3
            & 67.7 & 65.6 & 62.9 & 58.1 & 45.3
            & 67.7 & 65.6 & 62.9 & 58.1 & 45.3
            & 67.7 & 65.6 & 62.9 & 58.1 & 45.3
            & 67.7 & 65.6 & 62.9 & 58.1 & 45.3
            & 59.9 \\
        TENT-continual \cite{wang2020tent}
            & 66.8 & 64.8 & 62.3 & 58.1 & 45.6
            & 65.8 & 63.3 & 60.7 & 56.5 & 44.5
            & 64.2 & 61.3 & 58.6 & 54.6 & 43.1
            & 62.6 & 59.5 & 56.7 & 53.0 & 42.0
            & 61.1 & 58.1 & 55.4 & 51.8 & 41.2
            & 56.5 \\
        CoTTA \cite{wang2022continual}
            & 68.9 & 67.6 & 66.8 & 65.6 & 59.2
            & 69.2 & 68.1 & 67.2 & 66.0 & 60.3
            & 68.7 & 67.5 & 66.6 & 65.6 & 60.1
            & 68.6 & 67.5 & 66.6 & 65.8 & 60.3
            & 68.6 & 67.6 & 66.7 & 65.7 & 60.7
            & 65.8 \\
        SVDP \cite{yang2024exploring}
            & 69.2 & 68.1 & 66.9 & 65.5 & 61.7
            & 67.7 & 67.2 & 66.6 & 64.7 & 61.2
            & 67.5 & 66.8 & 66.1 & 65.3 & 62.6
            & 67.7 & 67.1 & 66.2 & 65.4 & 62.2
            & 67.4 & 66.6 & 66.1 & 65.4 & 62.1
            & 65.7 \\
        \midrule
        \textbf{Hybrid-TTA}
            & 68.2 & 67.7 & 67.4 & 66.8 & 62.9
            & 68.7 & 68.2 & 67.5 & 66.8 & 64.4
            & 68.1 & 67.7 & 67.1 & 66.5 & 64.5
            & 67.6 & 67.3 & 66.9 & 66.2 & 64.4
            & 67.4 & 67.1 & 66.7 & 66.0 & 64.3
            & \bf{66.7} \\
        \bottomrule
    \end{tabular}

\end{table}
\end{landscape}

\end{document}